%% file: neurips_2025.tex
\documentclass{article}

    \PassOptionsToPackage{numbers, compress}{natbib}

    \usepackage[final]{neurips_2025}

\usepackage[utf8]{inputenc} 
\usepackage[T1]{fontenc}    
\usepackage{url}            
\usepackage{booktabs}       
\usepackage{amsfonts}       
\usepackage{nicefrac}       
\usepackage{microtype}      
\usepackage[dvipsnames]{xcolor}
\usepackage[hidelinks]{hyperref}       

\usepackage{pifont}
\usepackage{dsfont}
\usepackage{xspace}
\usepackage{graphicx}
\usepackage{makecell}
\usepackage{multirow}
\usepackage{colortbl}
\usepackage{varwidth}
\usepackage{fontawesome}
\usepackage[listings]{tcolorbox}
\tcbuselibrary{breakable}
\usepackage{caption}
\usepackage{wrapfig}

\definecolor{MyRed}{HTML}{D2042D}
\definecolor{MyGreen}{HTML}{00A36C}
\definecolor{MyBlue}{HTML}{1F51FF}
\definecolor{MyPurple}{HTML}{AF58BA}
\definecolor{MyYellow}{HTML}{FFC61E}
\definecolor{MyOrange}{HTML}{F28522}
\definecolor{MyGrey}{HTML}{A0B1BA}
\definecolor{MyBrown}{HTML}{A6761D}
\definecolor{lightgray}{rgb}{0.83, 0.83, 0.83}
\definecolor{lavendermist}{rgb}{0.9, 0.9, 0.98}
\definecolor{lightblue}{HTML}{dfebf7}
\definecolor{lightgreen}{HTML}{f8fcf4}
\newcommand{\red}[1]{{\color{MyRed}#1}}
\newcommand{\green}[1]{{\color{MyGreen}#1}}
\newcommand{\blue}[1]{{\color{MyBlue}#1}}
\newcommand{\purple}[1]{{\color{MyPurple}#1}}

\newcommand{\grey}[1]{{\color{MyGrey}#1}}
\newcommand{\brown}[1]{{\color{MyBrown}#1}}
\newcommand{\bbox}{\purple{[bbox]}\xspace}
\newcommand{\bboxa}{\purple{[bbox1]}\xspace}
\newcommand{\bboxb}{\purple{[bbox2]}\xspace}

\newcommand{\model}{{\scshape SSR}\xspace}
\newcommand{\dataset}{{\scshape SSR-CoT}\xspace}
\newcommand{\benchmark}{{\scshape SSRBench}\xspace}

\title{\model: Enhancing Depth Perception in Vision-Language Models via Rationale-Guided Spatial Reasoning}

\author{
    Yang Liu$^{1,\dag}$, Ming Ma$^{3,\dag}$, Xiaomin Yu$^{4,\dag}$, Pengxiang Ding$^{1,2,\dag,\S}$, \\
    \textbf{Han Zhao$^{1,2}$, Mingyang Sun$^{1,2,5}$, Siteng Huang$^{2}$, Donglin Wang$^{1}$\thanks{Corresponding author. $^{\dag}$Equal contribution. $^{\S}$Project lead.}} \\
    $^{1}$Westlake University, $^{2}$Zhejiang University, $^{3}$Harbin Institute of Technology, \\
    $^{4}$The Hong Kong University of Science and Technology (Guangzhou), \\
    $^{5}$Shanghai Innovation Institute \\
    \texttt{\{liuyang67, wangdonglin\}@westlake.edu.cn} \\
}

\begin{document}

\maketitle

\newcommand{\px}[1]{\textcolor{blue}{#1}} 

\begin{abstract}
\input{sections/0_abs}
\end{abstract}

\input{sections/1_intro}
\input{sections/2_rel}
\input{sections/3_method}
\input{sections/4_exp}
\input{sections/5_ana}
\input{sections/6_con}

\begin{ack}
This work was supported by the National Science and Technology Innovation 2030 - Major Project (Grant No. 2022ZD0208800), and NSFC General Program (Grant No. 62176215).
\end{ack}

\newpage

\bibliographystyle{ieeetr}
\bibliography{neurips_2025}

\input{sections/7_appx}
\input{sections/8_checklist}

\end{document}

%% file: sections/0_abs.tex
Despite impressive advancements in Visual-Language Models (VLMs) for multi-modal tasks, their reliance on RGB inputs limits precise spatial understanding.
Existing methods for integrating spatial cues, such as point clouds or depth, either require specialized sensors or fail to effectively exploit depth information for higher-order reasoning.
To this end, we propose a novel Spatial Sense and Reasoning method, dubbed \model, a novel framework that transforms raw depth data into structured, interpretable textual rationales.
These textual rationales serve as meaningful intermediate representations to significantly enhance spatial reasoning capabilities.
Additionally, we leverage knowledge distillation to compress the generated rationales into compact latent embeddings, which facilitate resource-efficient and plug-and-play integration into existing VLMs without retraining.
To enable comprehensive evaluation, we introduce a new dataset named \dataset, a million-scale visual-language reasoning dataset enriched with intermediate spatial reasoning annotations, and present \benchmark, a comprehensive multi-task benchmark.
Extensive experiments on multiple benchmarks demonstrate \model substantially improves depth utilization and enhances spatial reasoning, thereby advancing VLMs toward more human-like multi-modal understanding.
Project page: \url{https://yliu-cs.github.io/SSR}.

%% file: sections/1_intro.tex
\section{Introduction}
\label{sec:intro}

\input{figures/teaser}

VLMs represent a pivotal advancement in bridging the gap between image and natural language, demonstrating astounding capabilities across myriad multi-modal tasks \cite{DBLP:conf/nips/LiuLWL23a, DBLP:conf/cvpr/LiuLLL24,DBLP:conf/cvpr/YuanLLTLQZZ24, DBLP:conf/cvpr/Rasheed0MS0CAX024, DBLP:conf/iccv/KirillovMRMRGXW23, conf/ecai/LiuPZSLZ23, DBLP:conf/eccv/LiuDHZZW24}.
Nevertheless, relying solely on RGB is inadequate for accurately capturing spatial information such as relative positions and distances, which presents inherent limitations in capturing precise spatial relationships, thereby constraining the capacity of VLMs to comprehend complex scenes.
Consequently, enhancing the ability of VLMs to understand and reason about spatial relationships is essential for critical real-world applications, particularly in robotics.

Recent advancements in VLMs have catalyzed research on explicitly incorporating spatial information to enhance model performance.
While some methods leverage point cloud data for improved spatial understanding \cite{DBLP:conf/icml/ZhenQCY0DHG24, DBLP:conf/nips/HongZCZDCG23, DBLP:conf/cvpr/0003XKISGX24}, they typically rely on specialized sensors (e.g., LiDAR) that are impractical in scenarios restricted to monocular RGB images.
In this context, monocular depth estimation has emerged as a compelling alternative, particularly with the proliferation of generative methods \cite{DBLP:conf/iclr/SongME21, DBLP:conf/nips/HoJA20}.
These methods enable the acquisition of high-quality depth images from standard 2D images through various pre-trained models \cite{DBLP:conf/cvpr/YangKHXFZ24, DBLP:conf/nips/YangKH0XFZ24, bochkovskiy2025depth, DBLP:journals/corr/abs-2502-19204}, eliminating additional hardware requirements.
By leveraging visual encoders pretrained on RGB images, depth features can be efficiently encoded and seamlessly integrated into VLMs, offering a promising pathway for enhancing spatial awareness.

However, a critical limitation of current methods lies in their superficial utilization of depth information \cite{DBLP:conf/icml/ZhenQCY0DHG24, DBLP:conf/nips/HongZCZDCG23, DBLP:conf/cvpr/0003XKISGX24, DBLP:journals/corr/abs-2406-13642, DBLP:conf/cvpr/ChenCZLY0Z0024, DBLP:conf/acl/ZhangCXL24, DBLP:journals/corr/abs-2406-10721}.
Unlike humans, who intuitively employ depth as an integral component within broader reasoning processes, existing methods incorporate depth explicitly without capitalizing on its inferential value \cite{DBLP:journals/corr/abs-2406-13642}.
Consider a query such as \textit{Are objects A and B far apart?}
Human cognition naturally analyzes the spatial relationship between objects and then leverages this understanding to inform subsequent reasoning.
This implicit reasoning process underscores the necessity for more sophisticated integration of depth information into VLMs, not merely as supplementary input, but as a fundamental component that facilitates complex spatial reasoning.
Developing methodologies that emulate this human-like implicit utilization of depth could substantially enhance VLM capabilities.

To this end, we propose \model, a novel paradigm designed to redefine the integration of depth information within VLMs.
Specifically, \model translates raw depth data into a structured \textit{rationale language}, providing an interpretable intermediate representation that bridges low-level depth perception and higher-level reasoning.
This rationale-based language facilitates VLMs in generating outputs that are both more accurate and contextually appropriate, while also enabling the previously underutilized inferential capabilities inherent to spatial depth information.
By converting modality-specific depth data into semantically rich and inherently aligned representations, \model effectively overcomes the interpretability limitations associated with traditional approaches.
Consequently, this method significantly enhances the utilization of depth information, laying the groundwork for more robust and human-like spatial reasoning capabilities within contemporary VLMs.
To further enhance the efficiency of rationale-language utilization, we transform depth information into a compact latent embedding.
Specifically, we apply a knowledge-distillation strategy to compress rationale-language representations into concise latent embeddings \cite{DBLP:conf/nips/LeeKPR24}.
Dissimilar to vanilla Chain-of-Thought (CoT) methods \cite{DBLP:conf/nips/Wei0SBIXCLZ22, DBLP:conf/nips/LeeKPR24, DBLP:journals/corr/abs-2412-16720, DBLP:journals/corr/abs-2501-12948} that rely primarily on textual explanations, our distillation strategy significantly reduces computational overhead while preserving the depth and inferential richness inherent to rationale-based representations \cite{DBLP:journals/corr/abs-2412-06769, DBLP:journals/corr/abs-2502-15589, feng2025efficient}.
Importantly, this module can be seamlessly integrated into existing VLMs via a training-free mechanism, highlighting the flexibility and broad applicability of the proposed framework.
To achieve \model, we first curate \dataset, a million-level vision-language spatial reasoning dataset that facilitates depth-aware reasoning and provides a robust foundation for developing sophisticated spatial reasoning models.
To validate our approach, we perform extensive experiments across multiple benchmarks.
Specifically, we also evaluate on our benchmark \benchmark, which comprises six distinct tasks spanning both general and spatial domains.
Extensive experiments and analysis demonstrate that \model substantially enhances spatial reasoning capabilities across diverse tasks, highlighting the effectiveness and broad utility of our proposed method.

Overall, our principal contributions in this paper are illustrated in Figure~\ref{fig:teaser} and summarized as follows: \\
$\bullet$ We propose an efficient VLM, dubbed \model, capable of simultaneously performing depth perception and spatial reasoning, and generating answers based on implicit reasoning rationales. \\
$\bullet$ We introduce \dataset, a million-scale visual-language reasoning dataset enriched with intermediate spatial reasoning annotations, and present \benchmark, a comprehensive multi-task benchmark. \\
$\bullet$ Extensive experiments and solid analysis across various benchmarks demonstrate our \model can efficiently and dramatically enhance the spatial understanding of existing VLMs.

%% file: figures/teaser.tex
\begin{figure}[!t]
    \centering
    \includegraphics[width=\linewidth]{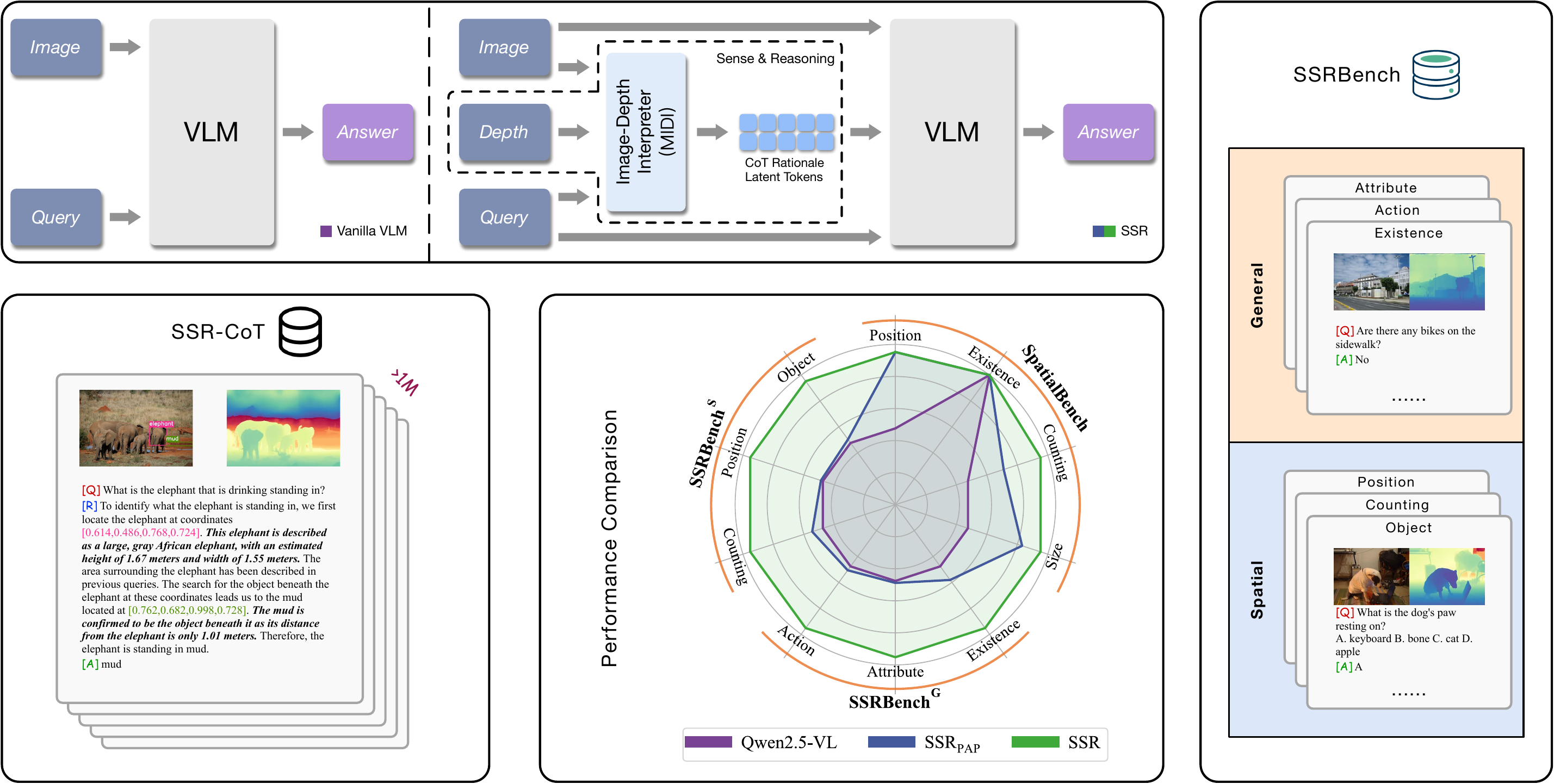}
    \caption{Unlike conventional VLMs, \model integrates depth perception to enhance spatial reasoning. We introduce a curated dataset \dataset and benchmark \benchmark, demonstrating significant improvements in spatial reasoning tasks.}
    \label{fig:teaser}
\end{figure}

%% file: sections/2_rel.tex
\section{Related Work}
\label{sec:rel}

\subsection{Visual-Language Models}

LLMs \cite{DBLP:journals/corr/abs-2501-12948, Radford2018ImprovingLU, Radford2019LanguageMA, conf/nips/BrownMRSKDNSSAA20, conf/nips/Ouyang0JAWMZASR22, OpenAI_ChatGPT, journals/corr/abs-2302-13971, journals/corr/abs-2307-09288, DBLP:journals/corr/abs-2309-16609, DBLP:journals/corr/abs-2407-10671, DBLP:journals/corr/abs-2407-21783, DBLP:journals/corr/abs-2412-15115} have led to major advancements in Natural Language Processing (NLP) tasks, and also have incited interest in developing VLMs.
Building a unified LLM with visual inputs for visual language tasks thus remains one of the most important desiderata for VLMs.
Over the last few years, VLMs achieved significant performance improvements in multi-modal tasks by integrating a pre-trained visual encoder and projecting the feature into semantic space into LLMs as well as training on large-scale multi-modal question-answering pairs \cite{DBLP:conf/nips/LiuLWL23a, DBLP:conf/cvpr/LiuLLL24, conf/nips/AlayracDLMBHLMM22, journals/corr/abs-2308-01390, DBLP:conf/cvpr/XuZYGA00S24, DBLP:conf/iclr/Zhu0SLE24, DBLP:conf/aaai/0008ZZDHW25}.
This straightforward method can work well for general tasks, but expecting the model to deduce answers for more complex tasks without deep reasoning can be daunting.

\subsection{Multi-Modal Reasoning}

LLMs display an emergent capability for step-by-step reasoning through in-context learning, a phenomenon referred to as CoT reasoning.
Such reasoning significantly enhances the performance of LLMs on complex reasoning tasks \cite{DBLP:conf/nips/Wei0SBIXCLZ22, DBLP:journals/corr/abs-2110-14168, DBLP:journals/tacl/GevaKSKRB21}.
Concurrently, notable advancements have also occurred in multi-modal CoT research, a paradigm appealing due to its similarity to human problem-solving behaviors \cite{DBLP:journals/corr/abs-2301-05226, DBLP:journals/tmlr/0001Z00KS24}.
Current research efforts in multi-modal CoT primarily emphasize the construction of intermediate reasoning rationale datasets to train image-text reasoning models.
Many existing studies adopt rich textual captions and detailed descriptions as intermediate rationales \cite{DBLP:journals/corr/abs-2404-09797, DBLP:conf/acml/JiaLLLG24, DBLP:conf/mm/GaoCZFSZZZSCJ24, DBLP:journals/corr/abs-2411-10440}.
Beyond text-based rationales, recent approaches have leveraged multi-modal rationales for more comprehensive reasoning \cite{DBLP:conf/nips/ShaoQ0SZW0024, DBLP:journals/corr/abs-2405-16919, DBLP:conf/nips/WuMZ000W24, DBLP:journals/corr/abs-2311-09241, DBLP:journals/corr/abs-2501-07542}.
However, existing multi-modal CoT methods primarily focus on tasks involving code generation, mathematical problem solving, and general question answering, which require sophisticated reasoning to achieve accurate responses.
In contrast, this paper introduces an efficient CoT method that leverages depth images to enhance the performance of VLMs, particularly by improving spatial understanding.

\subsection{Spatial Intelligence}

Spatial reasoning is an essential capability for VLMs and has therefore been included in several Visual Question Answering (VQA) benchmarks \cite{DBLP:conf/cvpr/HudsonM19, DBLP:conf/iclr/YiGLK0TT20, DBLP:journals/corr/abs-2411-16537, DBLP:journals/corr/abs-2412-14171}.
However, the majority of existing VLMs \cite{DBLP:conf/eccv/LiuDHZZW24, DBLP:conf/iclr/Chen0CPPSGGMB0P23, DBLP:conf/icml/0008LSH23, DBLP:conf/icml/DriessXSLCIWTVY23, DBLP:conf/icml/Wu0Q0C24} are primarily trained on two-dimensional images paired with textual data, a setting that inherently lacks comprehensive spatial information.
Consequently, these models exhibit limited performance in spatial reasoning tasks. To overcome this limitation, recent works such as SpatialVLM \cite{DBLP:conf/cvpr/0003XKISGX24}, SpatialRGPT \cite{DBLP:conf/nips/ChengYFGYK0L24} and RoboRefer \cite{DBLP:journals/corr/abs-2506-04308} have sought to improve the spatial reasoning capacity of VLMs by compiling specialized spatially-oriented question-answer datasets and fine-tuning models accordingly.
Nevertheless, despite these advancements, prior approaches largely neglect the integration of language-based reasoning capabilities within the spatial reasoning framework.
This omission hampers the effectiveness of existing VLMs in addressing more complex tasks, particularly those requiring intricate or multi-step reasoning processes.

%% file: sections/3_method.tex
\section{Methodology}
\label{sec:method}

\input{figures/arch}

\subsection{Architecture}

The primary goal of our proposed \model is to effectively leverage the reasoning capability of efficient language models effectively enhance the depth understanding and spatial reasoning capability for existing VLMs.
The overall framework is illustrated in Figure~\ref{fig:arch}.

\subsubsection{Image-Depth Interpreter}

To achieve comprehensive spatial understanding via depth interpretation, we propose a simple yet effective plug-and-play module entitled Mamba-based Image-Depth Interpreter (MIDI).
MIDI generates enriched depth-aware latent token representations, providing essential spatial reasoning information before feeding these tokens into the VLM.

Given an input image $X_{V}\in\mathbb{R}^{H\times W\times 3}$ and a corresponding textual query $X_{T}$, we first utilize a pretrained monocular depth estimation model, Depth Pro \cite{bochkovskiy2025depth}, to produce a depth $X_{D}\in\mathbb{R}^{H\times W\times 1}$ in the image.
Subsequently, image features $Z_{V}$ and depth features $Z_{D}$ are extracted from $X_{V}$ and $X_{D}$, respectively.
Specifically, we employ pre-trained CLIP ViT-L/14 \cite{DBLP:conf/icml/RadfordKHRGASAM21,DBLP:conf/iclr/DosovitskiyB0WZ21} as the visual encoder $\mathcal{E}_{V}$, and SigLIP \cite{DBLP:conf/iccv/ZhaiM0B23} as the depth encoder $\mathcal{E}_{D}$: $H_{\alpha} = \mathcal{E}_{\alpha}(X_{\alpha}), \alpha \in \{V, D\}$.
Then we apply Multi-Layer Perceptron (MLP) modules, comprising two fully connected layers with GELU \cite{DBLP:journals/corr/abs-1911-03925} activation, as projectors $\phi_{V}$ and $\phi_{D}$, transforming these visual features into the semantic embedding space compatible with the subsequent efficient language model: $Z_{\alpha} = \phi_{\alpha}(H_{\alpha}), \alpha \in \{V, D\}$.
To jointly encode visual and depth information conditioned on the textual query, we introduce an intermediate reasoning module implemented as the Mamba-based language model \cite{gu2024mamba}, denoted as $f_{\rm LM}$.
This module produces latent tokens representing intermediate spatial rationales: $H_{R} = f_{\rm LM}(Z_{V}, Z_{D}, X_{T})$.
Specifically, we uniformly insert several additional special tokens into the rationales to facilitate the knowledge distillation process and encode textual tokens into latent representations \cite{DBLP:conf/nips/LeeKPR24}.
Finally, similar to previous steps, we apply an additional latent projection module $\phi_{R}$ to map these latent rationale tokens into another semantic embedding space, matching the dimensionality of the word embeddings used in the subsequent VLM: $Z_{R} = \phi_{R}(H_{R})$.

Hence, our proposed MIDI module generates a sequence of spatial-aware latent tokens $Z_{R}$.
These tokens can easily be plugged into the query sequence for existing VLMs, effectively injecting depth-based spatial reasoning information and significantly enhancing the spatial understanding capabilities.

\input{tables/dataset}

\subsubsection{Spatial Sense and Reasoning}

Our proposed MIDI module fully leverages spatial information derived from the depth images and generates a latent representation $Z_{R}$, which encodes intermediate reasoning rationales essential for producing the response.
Subsequently, we input these latent tokens $Z_{R}$, alongside the original image $X_{V}$ and textual question $X_{T}$, into an existing VLM $f_{\rm VLM}$ to generate the answer $Y_{A}$ in an auto-regressive manner: $Y_{A} = f_{\rm VLM}(X_{V}, Z_{R}, X_{T})$.

\subsection{Training Paradigm}

For training the proposed \model, we adopt a two-stage procedure, as illustrated on the bottom-right side of Figure~\ref{fig:arch}.
In Stage 1, we train the underlying MIDI module to generate rationale latent tokens and project them into the language semantic space.
In Stage 2, we conduct joint training of the MIDI module and existing Vision-Language Models (VLMs) to further enhance performance.
Notably, Stage 2 is optional due to the modular and plug-and-play nature of our MIDI module, enabling straightforward integration into existing VLM frameworks.

\subsubsection{Stage 1: Reasoning and Alignment}

At the initial stage, we aim to train an efficient language model within the MIDI module, enabling it to generate and encode coherent thought processes represented by a sequence of features consistently aligned with the natural language semantic space.
To this end, each training sample at this phase includes a detailed and accurate rationale $Y_{R}$ as the ground truth.
After feeding latent tokens produced by the MIDI module into the subsequent LLM, we require the LLM to reconstruct the original textual rationale solely from these latent representations.
Training the LLM for precise rationale recovery depends not only on accurate reasoning capabilities of the MIDI module itself, but also on successfully projecting latent tokens into a semantic space consistent with the frozen-state LLM.

The learning objective for Stage 1 is defined by the standard causal modeling loss, given by:
\begin{equation}
\resizebox{0.8\linewidth}{!}{
    $\mathcal{L}_{1}(\theta) = -\mathbb{E}_{(X_V,X_D,X_T,Z_R,Y_R)\sim D}\left[\frac{1}{|Y_R|}\sum_{i=1}^{|Y_R|}\log P_{\theta}(Y_{R,i}\mid X_V,X_D,X_T,Z_R,Y_{R,<i})\right].$
}
\end{equation}

Following this training stage, latent tokens $Z_{R}$ generated by the MIDI module can be readily integrated into existing VLM image-text sequences, thereby enhancing their spatial understanding capabilities.

\subsubsection{Stage 2: Co-Training}

To further enhance the performance of \model, we jointly train the MIDI module along with existing VLMs.
Similar to instruction-tuning, we discard intermediate rationales in the second training stage and allow the VLM to directly generate the final answer.
In this setting, accurate answer generation by the VLM requires not only effective reasoning from the MIDI module but also the capacity of VLM to comprehend and utilize the reasoning information.
Specifically, the second-stage learning objective is formulated as a standard cross-entropy loss for auto-regressive generation of the final answer $Y_{A}$:
\begin{equation}
\resizebox{0.8\linewidth}{!}{
    $\mathcal{L}_{2}(\theta) = -\mathbb{E}_{(X_V,X_D,X_T,Y_A)\sim D}\left[\frac{1}{|Y_A|}\sum_{j=1}^{|Y_A|}\log P_{\theta}(Y_{A,j}\mid X_V,X_D,X_T,Y_{A,<j})\right].$
}
\end{equation}

\input{figures/data}

Since the rationale serving as the ground truth for supervised learning is omitted during Stage 2 training, we can incorporate additional VQA pairs to expand the dataset, thereby enhancing the generalization capability of the model.
Furthermore, training during this stage is optional due to the modular plug-and-play nature of the MIDI module.

%% file: figures/arch.tex
\begin{figure}[!t]
    \centering
    \includegraphics[width=\linewidth]{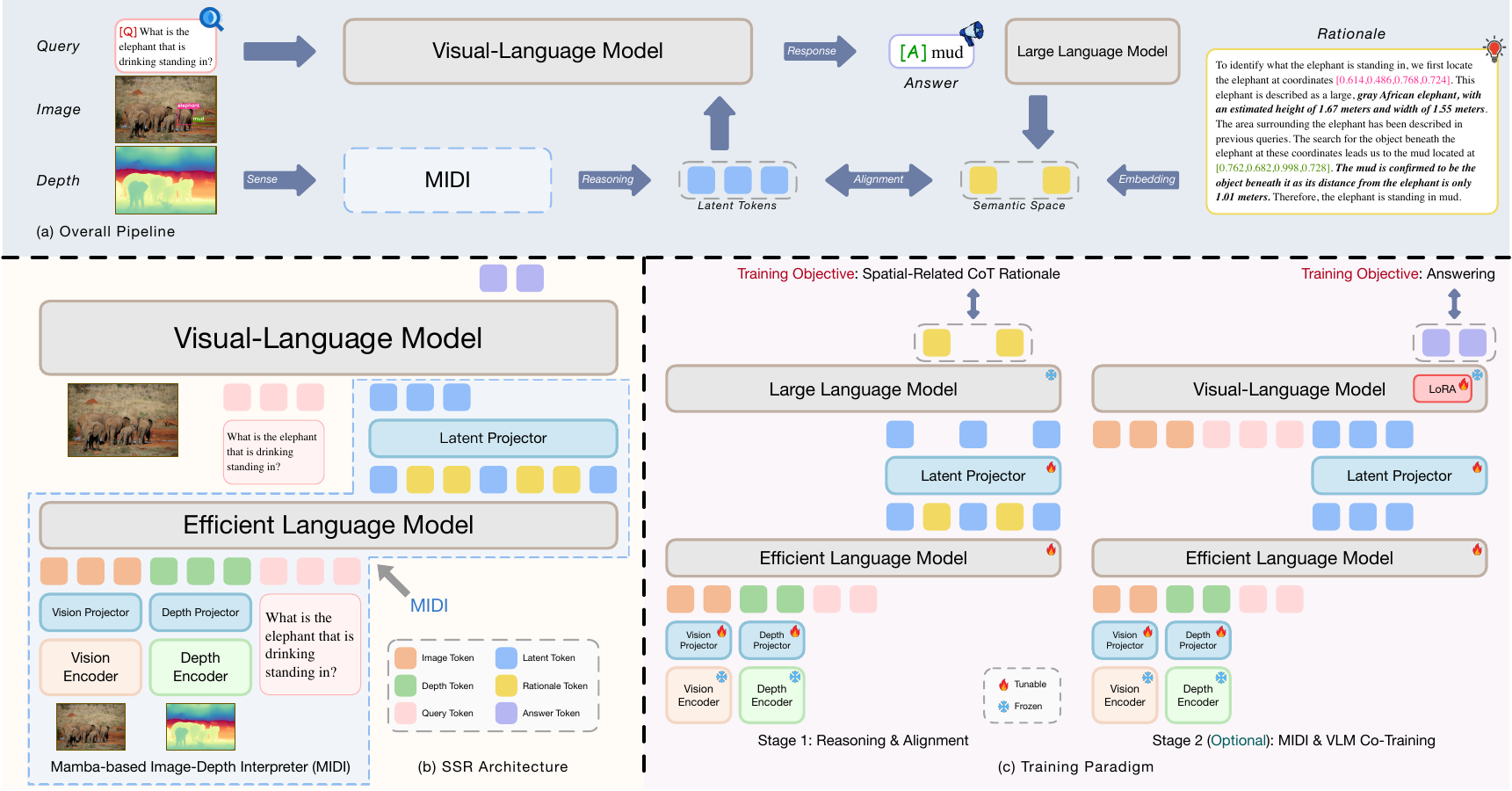}
    \caption{Schematic of \model framework. (a) Overall pipeline. (b) Full architecture of \model, comprising the MIDI module followed by the VLM. (c) Two training stages of the \model. In the stage 1, the LLM provides alignment supervision for the MIDI module, whereas the stage 2 is optional.}
    \label{fig:arch}
\end{figure}

%% file: tables/dataset.tex
\begin{table}
\caption{The mixture detail of \dataset dataset. \dataset consist over 1 million image-depth-question-rationale-answer pairs, where the rationale containing rich spatial-related knowledge the enhance Visual-Language Models (VLMs).}
\label{tab:dataset}
\centering
\resizebox{0.85\columnwidth}{!}{
\begin{tabular}{llllll}
\toprule
\multicolumn{1}{c}{{\bf Dataset}} & \multicolumn{1}{c}{{\bf Source}} & \multicolumn{1}{c}{{\bf Size}} & \multicolumn{1}{c}{{\bf Dataset}} & \multicolumn{1}{c}{{\bf Source}} & \multicolumn{1}{c}{{\bf Size}} \\
\midrule
\multirow{10}{*}{LLaVA-CoT \cite{DBLP:journals/corr/abs-2411-10440}} & ShareGPT4V \cite{DBLP:conf/eccv/ChenLDZHWZL24} & 31.3k & \multirow{6}{*}{Visual-CoT \cite{DBLP:conf/nips/ShaoQ0SZW0024}} & Flickr30k \cite{DBLP:conf/iccv/PlummerWCCHL15} & 136k\\
 & ChartQA \cite{DBLP:conf/acl/MasryLTJH22} & 17.2k & & GQA \cite{DBLP:conf/cvpr/HudsonM19} & 88k \\
 & A-OKVQA \cite{DBLP:conf/eccv/SchwenkKCMM22} & 16.1k & & Visual7W \cite{DBLP:conf/cvpr/ZhuGBF16} & 43k \\
 & AI2D \cite{DBLP:conf/eccv/KembhaviSKSHF16} & 11.4k & & OpenImages \cite{DBLP:journals/corr/abs-1811-00982} & 43k \\
 & GeoQA+ \cite{DBLP:conf/coling/CaoX22} & 11.4k & & Birds-200-2021 \cite{WahCUB_200_2011} & 10k \\
 & ScienceQA \cite{DBLP:conf/nips/LuMX0CZTCK22} & 5.6k & & VSR \cite{DBLP:journals/tacl/0001EC23} & 3k \\
\cmidrule{5-6}
 & DocVQA \cite{DBLP:conf/wacv/MathewKJ21} & 4.0k & \multicolumn{2}{r}{\cellcolor[rgb]{ .949,  .949,  .949}\textsc{Screened Total}} & \cellcolor[rgb]{ .949,  .949,  .949}\textit{289k} \\
\cmidrule{4-6}
 & PISC \cite{junnan_li_2017_1059155} & 1.0k & \multirow{3}{*}{VoCoT \cite{DBLP:journals/corr/abs-2405-16919}} & GQA \cite{DBLP:conf/cvpr/HudsonM19} & 72k \\
 & CLEVR \cite{DBLP:conf/cvpr/JohnsonHMFZG17} & 0.5k & & LLaVA-Instruct \cite{liu2024llavanext} & 6k \\
 & CLEVR-Math \cite{DBLP:conf/nesy/LindstromA22} & 0.5k & & LVIS \cite{DBLP:conf/cvpr/GuptaDG19} & 2k \\
\cmidrule{2-3} \cmidrule{5-6}
\multicolumn{2}{r}{\cellcolor[rgb]{ .949,  .949,  .949}\textsc{Total}} & \cellcolor[rgb]{ .949,  .949,  .949}\textit{98k} & \multicolumn{2}{r}{\cellcolor[rgb]{ .949,  .949,  .949}\textsc{Screened One-Turn Total}} & \cellcolor[rgb]{ .949,  .949,  .949}\textit{317k} \\
\midrule
SpatialQA \cite{DBLP:journals/corr/abs-2406-13642} & Bunny \cite{DBLP:journals/corr/abs-2402-11530} & 695k & & OXE \cite{DBLP:conf/icra/ONeillRMGPLPGMJ24} & 7.5k \\
\cmidrule{2-6}
\multicolumn{5}{r}{\cellcolor[rgb]{ .949,  .949,  .949}\textsc{Screened Total}} & \cellcolor[rgb]{ .949,  .949,  .949}\textit{501k} \\
\bottomrule
\end{tabular}
}
\end{table}

%% file: figures/data.tex

\begin{figure}[!t]
    \centering
    \begin{minipage}[c]{0.6\textwidth}
        \centering
        \includegraphics[width=\linewidth]{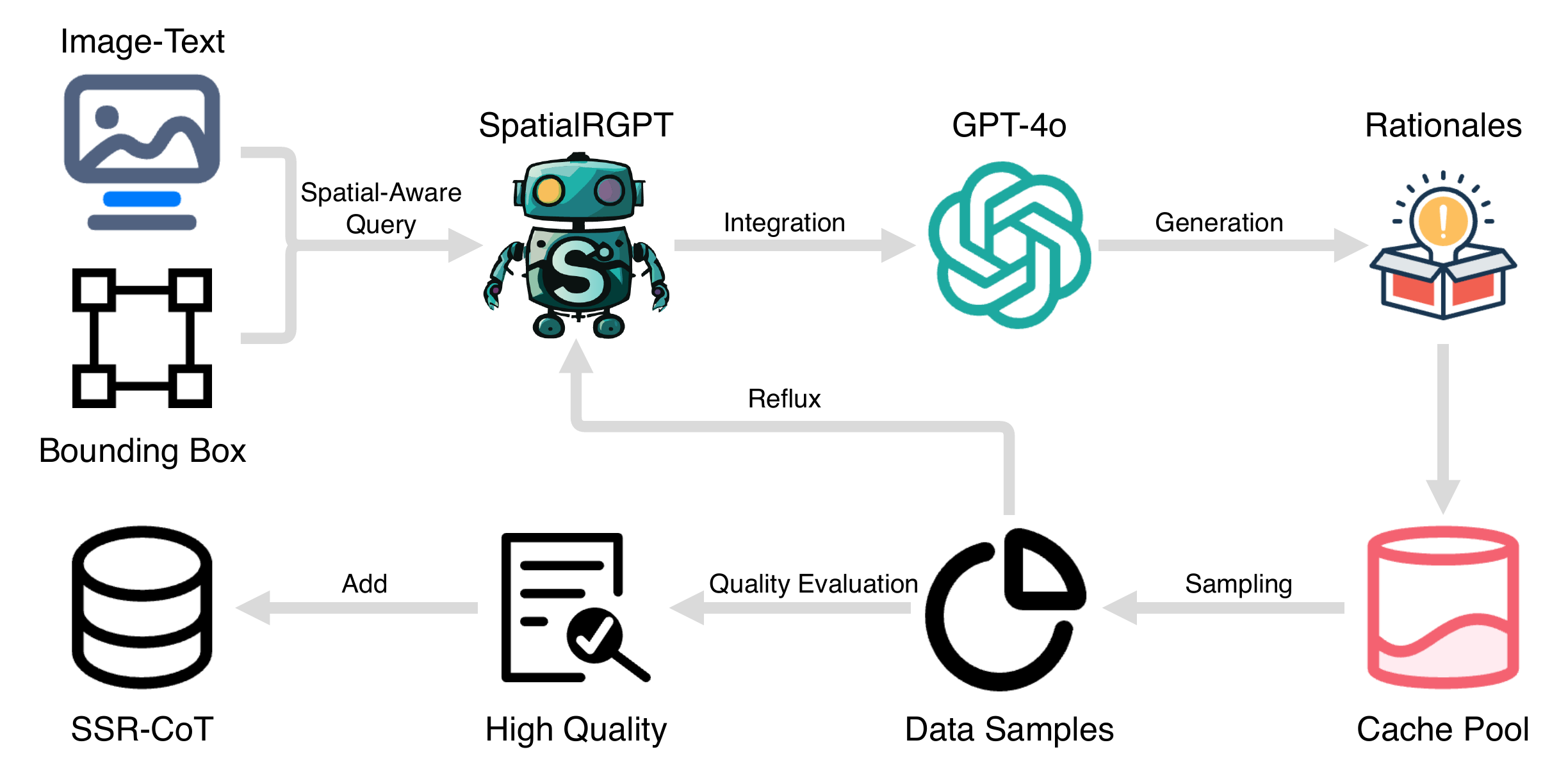}
        \caption{Schematic of SSR-CoT annotation pipeline.}
        \label{fig:data}
    \end{minipage}\hfill
    \begin{minipage}[c]{0.36\textwidth}
        \centering
        \captionof{table}{Quality evaluation for SSR-CoT dataset. We conduct the evaluation based on the powerful visual-language model Qwen2.5-VL-7B-Instruct \cite{bai2025qwen25vltechnicalreport}.}
        \label{tab:data_quality}
        \resizebox{\linewidth}{!}{
        \begin{tabular}{ccc}
            \toprule
            {\bf Rationale} & {\bf Accuracy} & {\bf Score} \\
            \midrule
            \red{\ding{55}} & 67.80 & 3.6721 \\
            {\cellcolor[rgb]{ .949,  .949,  .949}\green{\ding{51}}} & {\cellcolor[rgb]{ .949,  .949,  .949}\begin{tabular}[c]{@{}c@{}}79.42\\(\purple{$\uparrow$11.62})\end{tabular}} & {\cellcolor[rgb]{ .949,  .949,  .949}\begin{tabular}[c]{@{}c@{}}4.1289\\(\purple{$\uparrow$0.4568})\end{tabular}} \\
            \bottomrule
        \end{tabular}}
    \end{minipage}
\end{figure}

%% file: sections/4_exp.tex
\section{Experimentation}
\label{sec:exp}





\subsection{\dataset Collection}

There is a scarcity of visual-language CoT datasets with detailed reasoning processes annotations to train the \model model for depth perception and spatial understanding.
Therefore, we curate a new dataset from existing VQA datasets, resulting in over a total of 1 million image-depth-question-rationale-answer pairs.
There are four dataset sources we integrated:
(1) \textbf{LLaVA-CoT} \cite{DBLP:journals/corr/abs-2411-10440}: Systematic and structured reasoning visual-language CoT dataset, including general and science-targeted VQA data source.
(2) \textbf{Visual-CoT} \cite{DBLP:conf/nips/ShaoQ0SZW0024}: Multimodal CoT dataset that takes the bounding box as an intermediate thinking step, including general, relation reasoning and fine-grained science-targeted VQA data source.
(3) \textbf{VoCoT} \cite{DBLP:journals/corr/abs-2405-16919}: Fine-grained image-text CoT dataset that rationale provides detailed relationships between various objects with bounding box, including general and relation reasoning VQA data source.
(4) \textbf{SpatialQA} \cite{DBLP:journals/corr/abs-2406-13642}: Spatial QA dataset for sufficient utilization, including depth-related and robotic-related VQA data sources.

\input{figures/data_sample}

To generate visual-language reasoning data enriched with spatial information, we follow a multi-step process, as shown in Figure~\ref{fig:data}.
First, we extract depth estimations from raw images using Depth Pro \cite{bochkovskiy2025depth}.
For the LLaVA-CoT \cite{DBLP:journals/corr/abs-2411-10440} source, this is the only preprocessing step performed.
Second, for datasets such as VoCoT \cite{DBLP:journals/corr/abs-2405-16919} and SpatialQA \cite{DBLP:journals/corr/abs-2406-13642}, we refine long-form conversations by extracting concise, one-turn question-answer pairs.
Third, we leverage SpatialRGPT \cite{DBLP:conf/nips/ChengYFGYK0L24} to comprehensively mine precise spatial attributes within images, such as object size, distance, and relative positioning, based on intermediate reasoning steps, including bounding box annotations from Visual-CoT \cite{DBLP:conf/nips/ShaoQ0SZW0024} and VoCoT \cite{DBLP:journals/corr/abs-2405-16919}.
Finally, we employ GPT-4o \cite{DBLP:journals/corr/abs-2410-21276} to integrate all extracted information, generating detailed reasoning processes that enhance spatial understanding.
Notably, we also incorporate cache pools and perform sampling quality checks within iterative loops to ensure the high quality of the generated data.
Specifically, similar to the quality-assessment protocol shown in Table~\ref{tab:dataset}, we randomly draw 10\% of the cached samples and evaluate VQA accuracy both with and without their generated rationales. Rationales that improve accuracy are retained and incorporated into the final \dataset dataset; those that degrade performance are discarded, and the all samples in cache are re-submitted for re-annotation.
Overall, we compile approximately 1.2 million preprocessed data samples into \dataset dataset.
Figure~\ref{fig:data_sample} illustrates several data samples from the \dataset dataset.
Each data instance within \dataset comprises the original image, an associated question-answer pair, the corresponding estimated depth information, and a rationale.
The rationale incorporates fundamental reasoning steps used in question-answering tasks and provides detailed spatial reasoning to support accurate answer generation.

\input{figures/bench_sample}


To evaluate the quality of \dataset, we conducted an assessment based on the performance of the Qwen2.5-VL-7B-Instruct \cite{bai2025qwen25vltechnicalreport} on the VQA task.
This evaluation was carried out on a randomly selected subset comprising approximately 1\% of the full dataset, corresponding roughly to 10k samples.
Performance metrics include accuracy as well as a quantitative score ranging from 0 to 5, both are produced using the LLM-Assistant powered by the Qwen2.5-14B-Instruct-1M \cite{DBLP:journals/corr/abs-2412-15115, DBLP:journals/corr/abs-2501-15383}.
Further methodological details regarding the evaluation process are described in Appendix~\ref{sec:llm_eval}.
As presented in Table~\ref{tab:data_quality}, responses generated with intermediate rationales demonstrate an accuracy improvement of more than 10\% compared to direct question-answering.
This finding indicates that the intermediate reasoning rationales annotated in our dataset are of high quality and effectively enhance the question-answering performance of VLMs.

\subsection{\benchmark Construction}

Currently, there are no established benchmarks specifically designed for evaluating spatial understanding and reasoning capabilities on image-text pairs.
To address this gap, we propose \benchmark, a novel evaluation benchmark created from the \dataset dataset.
Importantly, the data incorporated into \benchmark will be fully removed from \dataset to prevent overlap.
\benchmark consists of two primary categories, general understanding and spatial understanding, allowing simultaneous evaluation of VLM performance in both general question answering and spatial reasoning tasks.
Each category contains three distinct evaluation tasks, with detailed sample sizes provided in Appendix~\ref{sec:bench}.

We illustrate the process to construct \benchmark as shown in Appendix~\ref{sec:bench_con}.
First, we define 6 distinct task categories.
Then, we randomly sampled image-text pairs from \dataset, proportionally retaining the distribution of its original data sources.
These samples were independently classified into task categories by GPT-4o \cite{DBLP:journals/corr/abs-2410-21276} and Gemini-2.5-Pro \cite{DBLP:journals/corr/abs-2403-05530}.
Only instances for which both models agreed on the assigned category were included in \benchmark; instances with disagreement were returned to \dataset.

In recent years, LLMs have demonstrated significant advancements in language understanding, reasoning, and text generation, exhibiting strong perceptual and comprehension capabilities through the implicit world knowledge they encapsulate.
Therefore, LLMs have increasingly been used as assessors to evaluate generation performance in question-answering tasks \cite{DBLP:conf/eccv/LiuDHZZW24, DBLP:conf/nips/LiuLWL23a, DBLP:journals/corr/abs-2503-07265, DBLP:conf/icml/YuYLWL0WW24}.
Consistent with our approach in data quality evaluation, we employ the Qwen2.5-14B-Instruct-1M \cite{DBLP:journals/corr/abs-2412-15115, DBLP:journals/corr/abs-2501-15383}, a powerful LLM, to evaluate the performance of VLMs in this benchmark.

\subsection{Implementation Details}

In this paper, we utilize Mamba \cite{gu2024mamba} as the lower-level efficient language model for reasoning, Qwen2.5 \cite{DBLP:journals/corr/abs-2412-15115} as the LLM for alignment in the first training stage, and Qwen2.5-VL \cite{bai2025qwen25vltechnicalreport} as the VLM supporting multi-modal comprehension in the second training stage.
During Stage 1, we exclusively train the MIDI component on our proposed \dataset dataset.
In Stage 2, we jointly train the \model using both the \dataset dataset and the LLaVA-Instruct-150K dataset \cite{DBLP:conf/nips/LiuLWL23a}.
Leveraging the efficiency of LoRA \cite{DBLP:conf/iclr/HuSWALWWC22} and Fully Sharded Data Parallel (FSDP) \cite{DBLP:journals/pvldb/ZhaoGVLHXWSOSDB23}, training \model requires approximately 19 hours for Stage 1 and 48 hours for Stage 2, using a single Nvidia 8-H800 GPU node equipped with 80GB VRAM.
Detailed hyperparameter configurations are provided in Appendix~\ref{sec:param}.

\input{tables/main}

\input{tables/allbench}

\subsection{Main Results}

Table~\ref{tab:main} presents the comparative performance of the \model against its backbone and state-of-the-art baselines on SpatialBench \cite{DBLP:journals/corr/abs-2406-13642} and \benchmark.
As shown by the results, our \model in 3 billion parameters can achieve comparable or even higher results than large-scale baseline models, including closed-source and backbone models. Our larger variant, comprising 7 billion parameters, yields the best performance on most tasks across the two benchmarks.
Compared to the top-performing baselines in each benchmark, \model exhibited notable improvements in the average question answering accuracy, achieving a maximum enhancement of 13.6 and an average improvement of 6.77.
Moreover, we also provide a detailed analysis of the performance improvements compared to the backbone model across additional benchmarks in Section~\ref{sec:perf_improv}.

%% file: figures/data_sample.tex
\begin{figure}[!t]
    \centering
    \includegraphics[width=\linewidth]{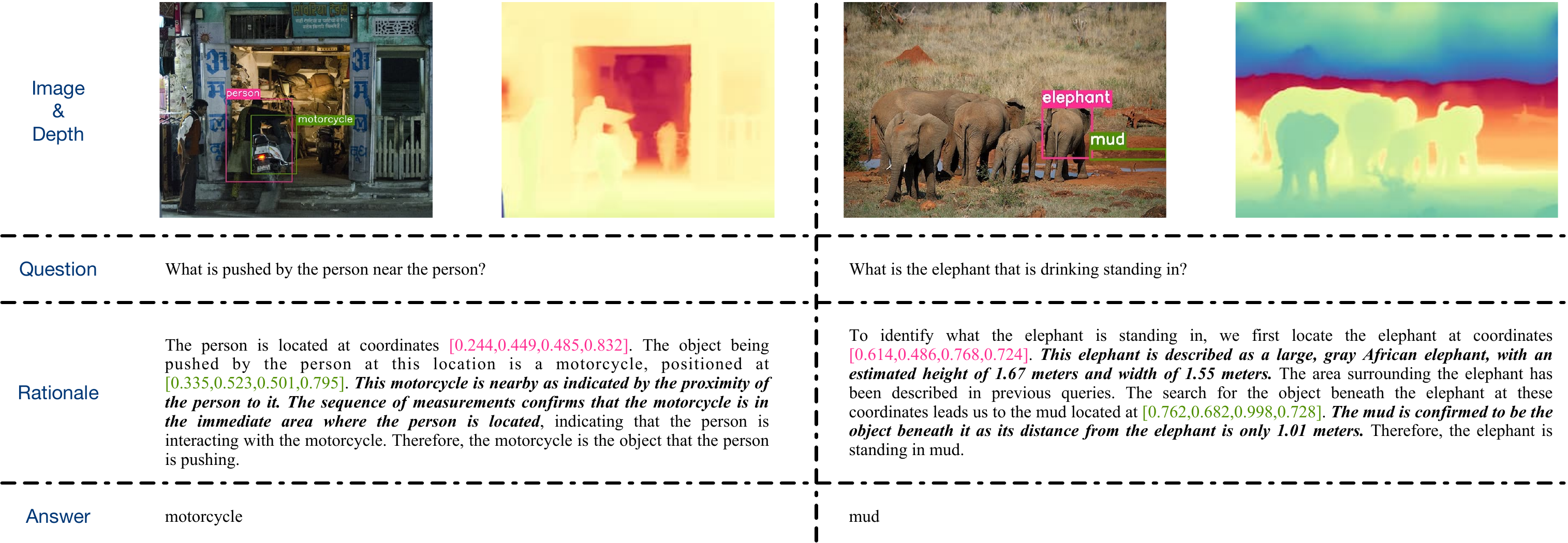}
    \caption{Illustrative samples of \dataset dataset.}
    \label{fig:data_sample}
\end{figure}

%% file: figures/bench_sample.tex
\begin{figure}[!t]
    \centering
    \includegraphics[width=\linewidth]{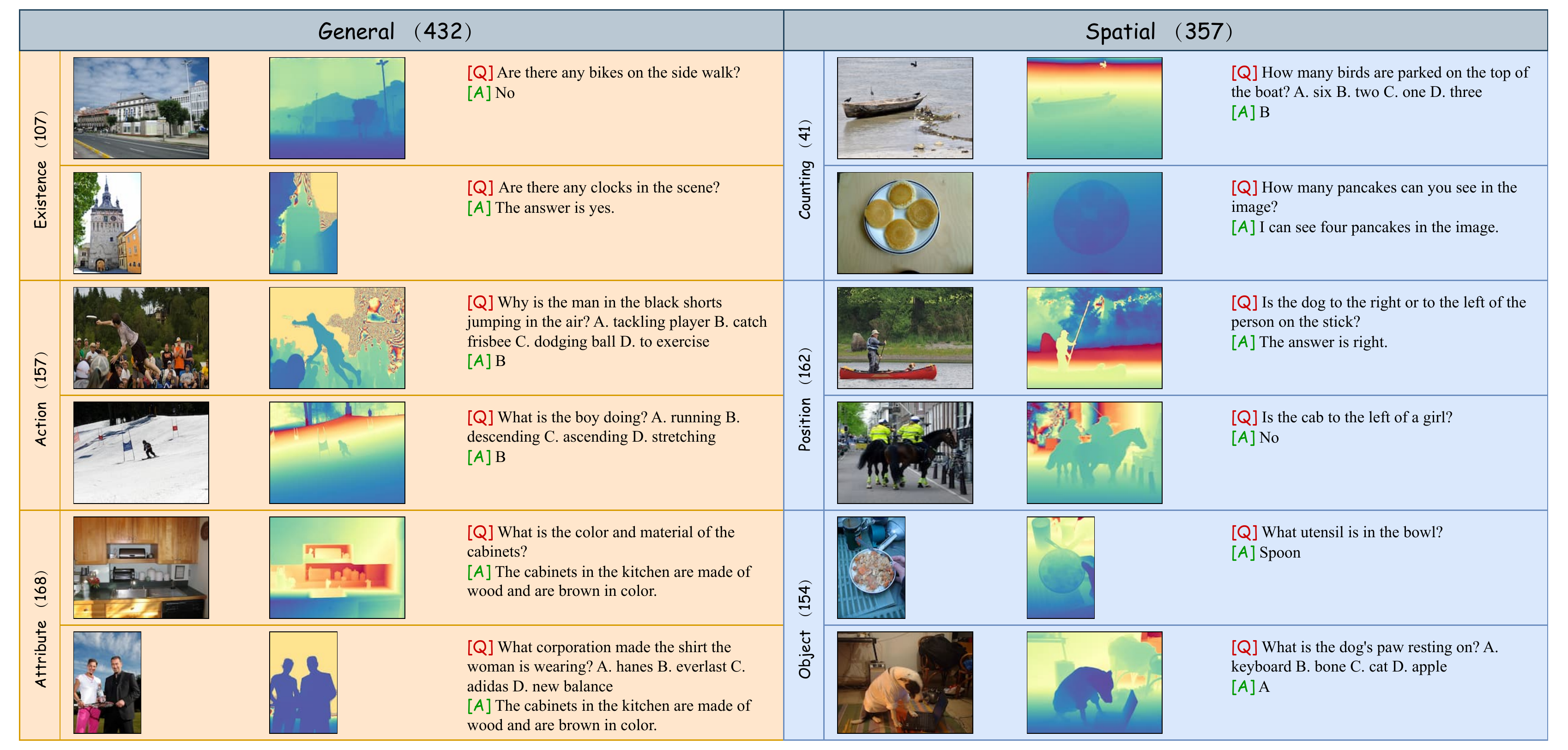}
    \caption{Examples for each task within the benchmark \benchmark.}
    \label{fig:bench_sample}
\end{figure}

%% file: tables/main.tex
\begin{table}[!t]
\caption{Performance comparison on SpatialBench \cite{DBLP:journals/corr/abs-2406-13642} and our \benchmark. \benchmark$^{\rm G}$ and \benchmark$^{\rm S}$ denote general and spatial tasks, respectively.}
\label{tab:main}
\centering
\resizebox{\columnwidth}{!}{
\begin{tabular}{lcccccccccccc} 
\toprule
\multicolumn{1}{c}{\multirow{2}{*}{\bf Method}} & \multicolumn{2}{c}{\bf Size} & \multicolumn{4}{c}{\bf SpatialBench \cite{DBLP:journals/corr/abs-2406-13642}} & \multicolumn{3}{c}{\bf \benchmark$^{\rm G}$} & \multicolumn{3}{c}{\bf \benchmark$^{\rm S}$} \\
\cmidrule(lr){2-3} \cmidrule(lr){4-7} \cmidrule(lr){8-10} \cmidrule(lr){11-13}
 & {\bf Mamba} & {\bf VLM} & \multicolumn{1}{c}{\bf Position} & \multicolumn{1}{c}{\bf Existence} & \multicolumn{1}{c}{\bf Counting} & \multicolumn{1}{c}{\bf Size} & \multicolumn{1}{c}{\bf Existence} & \multicolumn{1}{c}{\bf Attribute} & \multicolumn{1}{c}{\bf Action} & \multicolumn{1}{c}{\bf Counting} & \multicolumn{1}{c}{\bf Position} & \multicolumn{1}{c}{\bf Object} \\
\midrule
{\cellcolor[rgb]{ .949,  .949,  .949}\quad\quad \scshape Proprietary} & \multicolumn{12}{c}{\cellcolor[rgb]{ .949,  .949,  .949}} \\
GPT-4o-mini \cite{DBLP:journals/corr/abs-2410-21276} & \grey{N/A} & \grey{N/A} & 47.1 & 75.0 & 70.5 & 21.7 & 72.0 & 48.2 & 63.1 & 53.7 & 44.4 & 46.1 \\
Claude-3.5-Haiku \cite{anthropic_claude_3_5_haiku} & \grey{N/A} & \grey{N/A} & 55.9 & 65.0 & 72.2 & 26.7 & 51.4 & 52.2 & 43.7 & 42.0 & 34.1 & 38.3 \\
\midrule
{\cellcolor[rgb]{ .949,  .949,  .949}\quad\quad \scshape Open-Source} & \multicolumn{12}{c}{\cellcolor[rgb]{ .949,  .949,  .949}} \\
SpatialVLM \cite{DBLP:conf/cvpr/0003XKISGX24, VQASynth} & \grey{N/A} & 3B & 52.9 & 80.0 & 77.1 & 28.3 & 31.7 & 58.3 & 63.7 & 31.7 & 55.8 & 65.4 \\
LLaVA-1.5 \cite{DBLP:conf/cvpr/LiuLLL24} & \grey{N/A} & 7B & 44.1 & 45.0 & 82.8 & 30.0 & 81.3 & 64.3 & 66.9 & 43.9 & 63.6 & 63.6 \\
LLaVA-NeXT \cite{liu2024llavanext} & \grey{N/A} & 7B & 47.1 & 75.0 & 84.0 & 20.0 & 83.2 & 66.7 & 69.4 & 51.2 & 69.8 & 64.9 \\
LLaVA-NeXT \cite{liu2024llavanext} & \grey{N/A} & 13B & 47.1 & 75.0 & 82.9 & 20.0 & 86.9 & 69.6 & 71.3 & 41.5 & 69.8 & 53.2 \\
SpatialBot \cite{DBLP:journals/corr/abs-2406-13642} & \grey{N/A} & 3B & 50.0 & 80.0 & 86.7 & 25.0 & 75.7 & 61.3 & 67.5 & 39.0 & 74.1 & 61.7 \\
Emu3 \cite{DBLP:journals/corr/abs-2409-18869} & \grey{N/A} & 8B & 47.1 & 20.0 & 10.0 & 25.0 & 58.9 & 35.7 & 37.6 & 19.5 & 51.9 & 37.0 \\
Qwen2.5-VL \cite{bai2025qwen25vltechnicalreport} & \grey{N/A} & 3B & 55.9 & 80.0 & 76.4 & 25.0 & 66.4 & 58.9 & 63.1 & 34.1 & 60.5 & 51.9 \\
Qwen2.5-VL \cite{bai2025qwen25vltechnicalreport} & \grey{N/A} & 7B & 61.8 & 80.0 & 87.1 & 30.0 & 75.7 & 62.5 & 70.1 & 43.9 & 61.7 & 55.2 \\
\rowcolor{lightblue}{\bf \model (Ours)} & 130M & 3B & {\bf 64.7} & 80.0 & 82.9 & {\bf 31.7} & 83.2 & {\bf 82.1} & 72.6 & 51.2 & 83.3 & 74.7 \\
\rowcolor{lightblue}{\bf \model (Ours)} & 130M & 7B & {\bf 64.7} & {\bf 85.0} & {\bf 90.2} & 28.3 & {\bf 90.7} & 79.2 & {\bf 76.4} & {\bf 65.9} & {\bf 84.6} & {\bf 77.9} \\
\bottomrule
\end{tabular}
}
\end{table}

%% file: tables/allbench.tex
\begin{table}[!t]
\caption{Performance improvement of \model compared to the backbone model. SpatialBench \cite{DBLP:journals/corr/abs-2406-13642}, \benchmark, and CV-Bench \cite{DBLP:conf/nips/TongBWWIAYYMWPF24} report average.}
\label{tab:allbench}
\centering
\resizebox{\columnwidth}{!}{
\begin{tabular}{lcclllllll}
\toprule
\multicolumn{1}{c}{\multirow{2}{*}{\bf Method}} & \multicolumn{2}{c}{\bf Size} & \multicolumn{1}{c}{\multirow{2}{*}{\bf SpatialBench \cite{DBLP:journals/corr/abs-2406-13642}}} & \multicolumn{2}{c}{\bf \benchmark} & \multicolumn{1}{c}{\multirow{2}{*}{\bf CV-Bench \cite{DBLP:conf/nips/TongBWWIAYYMWPF24}}} & \multicolumn{2}{c}{\bf VSR \cite{DBLP:journals/tacl/0001EC23}} & \multirow{2}{*}{\bf What's Up \cite{DBLP:conf/emnlp/KamathHC23a}} \\
\cmidrule(lr){2-3} \cmidrule(lr){5-6} \cmidrule(lr){8-9}
 & {\bf Mamba} & {\bf VLM} & & \multicolumn{1}{c}{\bf General} & \multicolumn{1}{c}{\bf Spatial} & & \multicolumn{1}{c}{\bf Random} & \multicolumn{1}{c}{\bf Zero-Shot} \\
\midrule
Qwen2.5-VL \cite{bai2025qwen25vltechnicalreport} & \grey{N/A} & 3B & 59.3 & 62.8 & 48.8 & 67.0 & 73.0 & 76.4 & 85.4 \\
\rowcolor{lightblue}{\bf \quad \model (Ours)} & 130M & 3B & 64.8 (\purple{$\uparrow$5.4}) & 79.3 (\purple{$\uparrow$16.5}) & 69.7 (\purple{$\uparrow$20.9}) & 68.9 (\purple{$\uparrow$1.9}) & 78.6 (\purple{$\uparrow$5.6}) & 82.9 (\purple{$\uparrow$6.5}) & 87.9 (\purple{$\uparrow$2.5}) \\
Qwen2.5-VL \cite{bai2025qwen25vltechnicalreport} & \grey{N/A} & 7B & 64.7 & 69.4 & 53.6 & 73.0 & \grey{N/A} & \grey{N/A} & \grey{N/A} \\
\rowcolor{lightblue}{\bf \quad \model (Ours)} & 130M & 7B & 67.0 (\purple{$\uparrow$2.3}) & 82.1 (\purple{$\uparrow$12.6}) & 76.1 (\purple{$\uparrow$22.5}) & 73.3 (\purple{$\uparrow$0.3}) & \grey{N/A} & \grey{N/A} & \grey{N/A} \\
\bottomrule
\end{tabular}
}
\end{table}

%% file: sections/5_ana.tex
\section{Analysis}
\label{sec:ana}


\subsection{Performance Improvement}
\label{sec:perf_improv}

We present additional experimental results in Table~\ref{tab:allbench}, demonstrating the improved performance of \model compared to the backbone model across the five benchmarks shown in Table~\ref{tab:bench} at varying model scales.
Specifically, across the three benchmarks reporting average values, SSR models of different sizes demonstrated average improvements of \textbf{11.2} and \textbf{9.4} compared to the backbone model.
The most significant improvements were observed in the space task of the benchmark, where the enhancements reached \textbf{20.9} and \textbf{22.5}, respectively.
This result exceeds the improvements reported in Table~\ref{tab:data_quality}, indicating that our \model effectively reasons about information highly relevant to multi-modal VQA tasks without introducing significant additional noise.
Furthermore, the training paradigm of \model enhances performance not only on two evaluation datasets closely related to the training data but also on the out-of-domain CV-Bench \cite{DBLP:conf/nips/TongBWWIAYYMWPF24}, VSR \cite{DBLP:journals/tacl/0001EC23}, What's Up \cite{DBLP:conf/emnlp/KamathHC23a} and multiple general VQA benchmarks \cite{DBLP:conf/cvpr/HudsonM19, DBLP:conf/cvpr/GoyalKSBP17, DBLP:conf/cvpr/SinghNSJCBPR19, DBLP:conf/emnlp/LiDZWZW23, DBLP:conf/eccv/LiuDZLZZYWHLCL24} as shown in Table~\ref{tab:general_vqa}.
These findings indicate that our training approach effectively further improves the generality and generalization capability of the \model in addition to enhanced spatial understanding performance.

\subsection{Ablation Studies}
\label{sec:abla}

As reported in Table~\ref{tab:abla}, these experiments illustrate the performance of the MIDI module when integrated in a plug-and-play manner without second training stage, leading to improved spatial understanding.
Specifically, this approach achieves average performance gains of \textbf{4.4} and \textbf{1.6} on different benchmark datasets.
On certain tasks, the plug-and-play approach achieves performance improvements of up to \textbf{8.8}, demonstrating the effectiveness of this usage.
In addition, after the second stage training, the performance of the complete \model model will be significantly improved on this basis, achieving average performance gains of \textbf{5.7} and \textbf{18.7} on different benchmark datasets.
Moreover, we provide case studies in Appendix~\ref{sec:case_study}.

\input{tables/general_vqa}

\subsection{Efficiency}
\label{sec:eff}

\input{tables/abla}

We evaluate the Qwen2.5-VL \cite{bai2025qwen25vltechnicalreport} after fine-tuning on \dataset dataset, overly protracted and convoluted textual intermediate reasoning chains not only increase the risk of erroneous conclusions but also impose prohibitive computational costs that undermine inference efficiency, which can better reflect the importance of latent reasoning method in CoT application.
As illustrated in Table~\ref{tab:efficiency}, the results demonstrate that, although \model introduces a modest absolute latency per generated token, its latent-reasoning paradigm dramatically curtails the number of CoT tokens needed to reach a final response. Consequently, under the CoT-based evaluation framework, the overall end-to-end inference speed is substantially improved.

\subsection{LLM-Assistant Evaluation}
\label{sec:llm_eval}

\input{tables/efficiency}

\input{tables/llm_eval}

To mitigate potential biases arising from employing models of the same family as judges, we re-assessed the \benchmark results with GPT-4o-mini \cite{DBLP:journals/corr/abs-2410-21276}, the outcomes are reported in Table~\ref{tab:llm_eval}.
High inter-model agreement between the evaluation scores assigned by different LLM judges in the table suggests that simple answer-comparison tasks largely resist bias within model series.

\subsection{Rationale Embedding}
\label{sec:vis}

\begin{wrapfigure}{r}{0.28\textwidth}
    \vspace{-40pt}
    \centering
    \includegraphics[width=\linewidth,trim=0 0 0 0,clip]{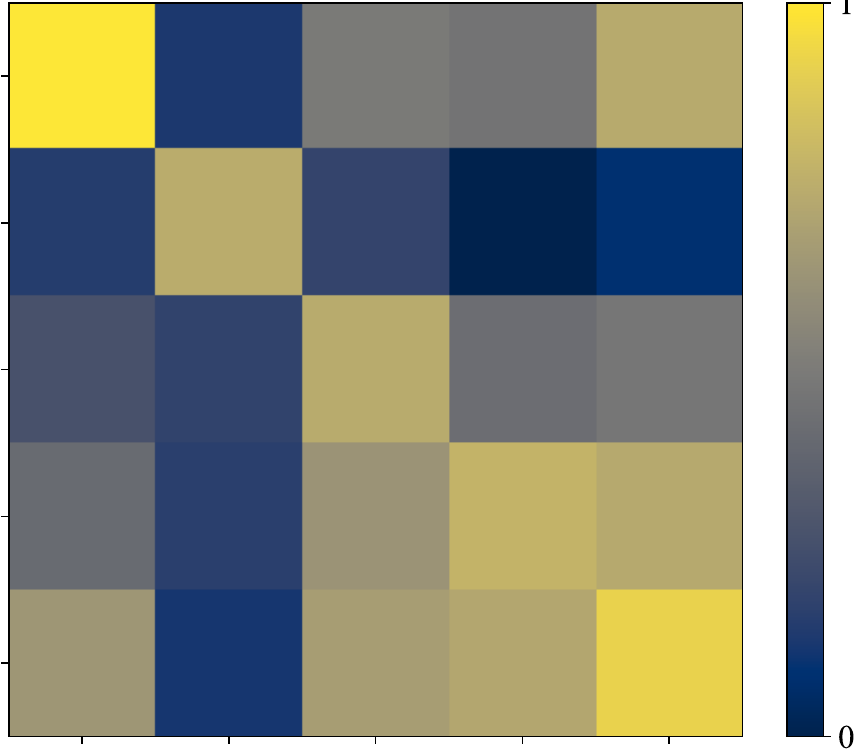}
    \caption{Cosine similarity matrix of reasoning latent tokens with/without rationale.}
    \label{fig:cos_sim}
    \vspace{-5pt}
\end{wrapfigure}

To analyze whether MIDI effectively captures depth information and conducts spatial reasoning guided by rationale, we visualize the cosine similarity between latent tokens, both with and without rationale.
Figure~\ref{fig:cos_sim} visualizes the cosine similarities between the latent tokens produced in two different paradigms: x-axis: latent tokens inserted inside the rationale, y-axis: latent tokens inserted immediately after the question and used to start the answer generation. Diagonal cells represent these two states of the same sample. High values on the diagonal indicate that the model has learned to map the rationale to the latent representation, confirming that it successfully distills the spatial knowledge embedded in the rationale. Low off-diagonal values indicate that the latent tokens remain sample-specific and do not collapse to a generic representation.

%% file: tables/general_vqa.tex
\begin{table}[!t]
\caption{Performance comparison on general VQA benchmarks}
\label{tab:general_vqa}
\centering
\resizebox{\columnwidth}{!}{
\begin{tabular}{lcclllll}
\toprule
\multicolumn{1}{c}{\multirow{2}{*}{\bf Method}} & \multicolumn{2}{c}{\bf Size} & \multicolumn{1}{c}{\multirow{2}{*}{\bf VQAv2 \cite{DBLP:conf/cvpr/GoyalKSBP17}}} & \multicolumn{1}{c}{\multirow{2}{*}{\bf TextVQA \cite{DBLP:conf/cvpr/SinghNSJCBPR19}}} & \multicolumn{1}{c}{\multirow{2}{*}{\bf POPE \cite{DBLP:conf/emnlp/LiDZWZW23}}} & \multicolumn{1}{c}{\multirow{2}{*}{\bf MMBench \cite{DBLP:conf/eccv/LiuDZLZZYWHLCL24}}} & \multirow{2}{*}{\bf GQA \cite{DBLP:conf/cvpr/HudsonM19}} \\
\cmidrule(lr){2-3}
 & {\bf Mamba} & {\bf VLM} & & & & \\
\midrule
Qwen2.5-VL \cite{bai2025qwen25vltechnicalreport} & \grey{N/A} & 3B & 72.5 & 57.0 & 84.4 & 75.9 & 56.2 \\
\rowcolor{lightblue}{\bf \quad \model (Ours)} & 130M & 3B & 79.0 (\purple{$\uparrow$6.5}) & 61.3 (\purple{$\uparrow$4.3}) & 86.0 (\purple{$\uparrow$1.6}) & 78.3 (\purple{$\uparrow$2.4}) & 63.6 (\purple{$\uparrow$7.4}) \\
\bottomrule
\end{tabular}
}
\end{table}

%% file: tables/abla.tex
\begin{table}[!t]
\caption{Performance comparison among the backbone model, the \model with/without the second training stage. \textsubscript{\scshape PaP} indicates that the MIDI module was employed in a plug-and-play manner.}
\label{tab:abla}
\centering
\resizebox{\columnwidth}{!}{
\begin{tabular}{lcccccccccccc}
\toprule
\multicolumn{1}{c}{\multirow{2}{*}{\bf Method}} & \multicolumn{2}{c}{\bf Size} & \multicolumn{4}{c}{\bf SpatialBench \cite{DBLP:journals/corr/abs-2406-13642}} & \multicolumn{3}{c}{\bf \benchmark$^{\rm G}$} & \multicolumn{3}{c}{\bf \benchmark$^{\rm S}$} \\
\cmidrule(lr){2-3} \cmidrule(lr){4-7} \cmidrule(lr){8-10} \cmidrule(lr){11-13}
 & {\bf Mamba} & {\bf VLM} & {\bf Position} & {\bf Existence} & {\bf Counting} & {\bf Size} & {\bf Existence} & {\bf Attribute} & {\bf Action} & {\bf Counting} & {\bf Position} & {\bf Object} \\
\midrule
Qwen2.5-VL \cite{bai2025qwen25vltechnicalreport} & \grey{N/A} & 3B & 55.9 & 80.0 & 76.4 & 25.0 & 66.4 & 58.9 & 63.1 & 34.1 & 60.5 & 51.9 \\
\cellcolor{lightgreen}{\quad {\bf \model}\textsubscript{\scshape PaP}} & \cellcolor{lightgreen}{130M} & \cellcolor{lightgreen}{3B} 
& \cellcolor{lightgreen}{\begin{tabular}[c]{@{}c@{}}64.7\\(\purple{$\uparrow$8.8})\end{tabular}} 
& \cellcolor{lightgreen}{\begin{tabular}[c]{@{}c@{}}80.0\\(\grey{0.0})\end{tabular}} 
& \cellcolor{lightgreen}{\begin{tabular}[c]{@{}c@{}}79.6\\(\purple{$\uparrow$3.2})\end{tabular}} 
& \cellcolor{lightgreen}{\begin{tabular}[c]{@{}c@{}}30.0\\(\purple{$\uparrow$5.0})\end{tabular}} 
& \cellcolor{lightgreen}{\begin{tabular}[c]{@{}c@{}}70.1\\(\purple{$\uparrow$3.7})\end{tabular}} 
& \cellcolor{lightgreen}{\begin{tabular}[c]{@{}c@{}}59.5\\(\purple{$\uparrow$0.6})\end{tabular}} 
& \cellcolor{lightgreen}{\begin{tabular}[c]{@{}c@{}}63.7\\(\purple{$\uparrow$0.6})\end{tabular}} 
& \cellcolor{lightgreen}{\begin{tabular}[c]{@{}c@{}}36.6\\(\purple{$\uparrow$2.5})\end{tabular}} 
& \cellcolor{lightgreen}{\begin{tabular}[c]{@{}c@{}}61.1\\(\purple{$\uparrow$0.6})\end{tabular}} 
& \cellcolor{lightgreen}{\begin{tabular}[c]{@{}c@{}}53.2\\(\purple{$\uparrow$1.3})\end{tabular}} \\
\cellcolor{lightblue}{\quad {\bf \model}} & \cellcolor{lightblue}{130M} & \cellcolor{lightblue}{3B} 
& \cellcolor{lightblue}{\begin{tabular}[c]{@{}c@{}}64.7\\(\purple{$\uparrow$8.8})\end{tabular}} 
& \cellcolor{lightblue}{\begin{tabular}[c]{@{}c@{}}80.0\\(\grey{0.0})\end{tabular}} 
& \cellcolor{lightblue}{\begin{tabular}[c]{@{}c@{}}82.9\\(\purple{$\uparrow$6.5})\end{tabular}} 
& \cellcolor{lightblue}{\begin{tabular}[c]{@{}c@{}}31.7\\(\purple{$\uparrow$6.7})\end{tabular}} 
& \cellcolor{lightblue}{\begin{tabular}[c]{@{}c@{}}83.2\\(\purple{$\uparrow$16.8})\end{tabular}} 
& \cellcolor{lightblue}{\begin{tabular}[c]{@{}c@{}}82.1\\(\purple{$\uparrow$23.2})\end{tabular}} 
& \cellcolor{lightblue}{\begin{tabular}[c]{@{}c@{}}72.6\\(\purple{$\uparrow$9.5})\end{tabular}} 
& \cellcolor{lightblue}{\begin{tabular}[c]{@{}c@{}}51.2\\(\purple{$\uparrow$17.1})\end{tabular}} 
& \cellcolor{lightblue}{\begin{tabular}[c]{@{}c@{}}83.3\\(\purple{$\uparrow$22.8})\end{tabular}} 
& \cellcolor{lightblue}{\begin{tabular}[c]{@{}c@{}}74.7\\(\purple{$\uparrow$22.8})\end{tabular}} \\
\bottomrule
\end{tabular}
}
\end{table}

%% file: tables/efficiency.tex
\begin{table}
\caption{Inference efficiency comparison on SpatialBench \cite{DBLP:journals/corr/abs-2406-13642}.}
\label{tab:efficiency}
\centering
\resizebox{\columnwidth}{!}{
\begin{tabular}{llllll}
\toprule
{\bf Model} & {\bf Size} & {\bf SpatialBench \cite{DBLP:journals/corr/abs-2406-13642}} & {\bf Token Per Sample} & {\bf Token Per Second} & {\bf Inference Time Per Sample} \\
\midrule
Qwen2.5-VL \cite{bai2025qwen25vltechnicalreport} (w/ SFT on \dataset) & 3B & 51.3 & 437.28 & 18.88 & 23.16s \\
{\bf \model (Ours)} & 3B & 64.8 & 2.62 & 8.18 & 0.32s \\
\bottomrule
\end{tabular}
}
\end{table}

%% file: tables/llm_eval.tex
\begin{table}
\caption{LLM-Assistant Evaluation Performance comparison on \benchmark, evaluated using different LLMs.  (For each result, the left column presents the original Qwen-based evaluation score, while the right column reports the corresponding GPT judgment).}
\label{tab:llm_eval}
\centering
\resizebox{\columnwidth}{!}{
\begin{tabular}{llllllll}
\toprule
{\bf Model} & {\bf Size} & {\bf Existence} & {\bf Attribute} & {\bf Action} & {\bf Counting} & {\bf Position} & {\bf Object} \\
\midrule
Qwen2.5-VL \cite{bai2025qwen25vltechnicalreport} & 3B & 66.4 / 58.9 & 58.9 / 59.5 & 63.1 / 67.5 & 34.1 / 34.1 & 60.5 / 58.6 & 51.9 / 52.6 \\
{\bf \model (Ours)} & 3B & 83.2 / 81.3 & 82.1 / 79.8 & 72.6 / 73.3 & 51.2 / 51.2 & 83.3 / 80.8 & 74.7 / 74.7 \\
\bottomrule
\end{tabular}
}
\end{table}

%% file: sections/6_con.tex
\section{Conclusion}
\label{sec:con}

In this paper, we propose a novel VLM \model with an important module named MIDI to interpret depth for enhancing the depth perception and spatial reasoning capabilities of existing VLMs.
MIDI can even be efficiently integrated into existing VLMs in a seamless, plug-and-play manner.
To enable effective training and evaluation, we curate a multi-modal CoT dataset \dataset and present a comprehensive benchmark \benchmark.
Extensive experiments conducted across four distinct benchmarks demonstrate that \model consistently achieves state-of-the-art performance enhancements over existing approaches, particularly excelling in spatially-oriented visual question answering tasks.

\paragraph{Broader Impacts.}
Our proposed \model demonstrates that spatial reasoning capabilities can be incrementally enhanced without adversely affecting its existing VLM functionalities.
This provides an innovative avenue for research communities to integrate additional capabilities into VLMs.

\paragraph{Limitation and Future Works.}
Although \model shows astounding performance, this study is limited to the Qwen/Qwen-VL series; future work will broaden the VLM scope to test generalizability.

%% file: sections/7_appx.tex
\newpage

\appendix
\label{section:appendix}

\section{Hyperparameters}
\label{sec:param}

Detailed hyperparameter configurations are provided in Table~\ref{tab:param}.

\input{tables/param}

\section{\benchmark Results in Score Metrics}
\label{sec:ssr_score}

The evaluation metrics employed for \benchmark include both accuracy and a quantitative score ranging from 0 to 5.
Quantitative results are presented in Table~\ref{tab:ssr_score}, with detailed descriptions of the assessment methodology provided in Appendix~\ref{sec:llm_eval}.
These scores are generally consistent with the accuracy trends presented in Table~\ref{tab:main}.

\input{tables/ssrbench_score}

\paragraph{Discussion with Meteor.}
Meteor \cite{DBLP:conf/nips/LeeKPR24} is an approach similar to ours, designed to compress rationales using efficient large language models.
However, unlike our method, Meteor does not separate the reasoning module from the large language model during response generation.
Due to this tight coupling, Meteor must be trained end-to-end from scratch, a process that demands extensive datasets and significant computational resources.
In contrast, our method specifically focuses on enhancing the spatial awareness and reasoning abilities of Vision-Language Models (VLMs), leveraging their inherent capabilities to a greater extent.
Consequently, our approach substantially reduces the complexity and resource requirements related to training VLMs from scratch.
Moreover, we focus on directionally enhancing the depth perception and spatial reasoning capabilities of existing VLMs in this paper.
Therefore, our comparative analysis primarily emphasizes evaluating model performance before and after applying these enhancements.

\section{\dataset}
\label{sec:data}

As detailed in Section~\ref{sec:exp}, the \dataset is constructed from four distinct data sources, with spatially-aware CoT rationales generated for each data sample.
Representative samples are shown in Figure~\ref{fig:data_sample}.
Additionally, a comprehensive description of the rationale-generation pipeline specific to each data source is presented in Appendix~\ref{sec:apx_vocot}, \ref{sec:apx_viscot}, and \ref{sec:apx_spaqa}.




\subsection{Visual-CoT}
\label{sec:apx_viscot}

For Visual-CoT \cite{DBLP:conf/nips/ShaoQ0SZW0024}, each data sample includes a bounding box that serves as a CoT rationale to guide the generation of the corresponding answer.
We utilize this bounding box, which is closely related to the target answer, as an intermediate step to query spatial information about the selected object using SpatialRGPT \cite{DBLP:conf/nips/ChengYFGYK0L24}, a spatial question-answering model tailored for vertical domains.
Subsequently, we aggregate the obtained spatial question-and-answer information using a powerful LLM such as GPT-4o \cite{DBLP:journals/corr/abs-2410-21276}.
The resulting text serves as the CoT rationale for the Visual-CoT data source within \dataset.

\input{boxes/viscot_qry}

\input{boxes/viscot_cmb}

\subsection{VoCoT}
\label{sec:apx_vocot}

VoCoT \cite{DBLP:journals/corr/abs-2405-16919} includes multiple bounding boxes per data sample, more than Visual-CoT \cite{DBLP:conf/nips/ShaoQ0SZW0024}, to clearly outline reasoning paths involving multiple objects within an image.
Similar to the process used for Visual-CoT, we perform spatial queries on each object associated with a bounding box.
Additionally, we capture the relative spatial relationships between every pair of objects to comprehensively utilize available spatial context and support accurate reasoning.
Finally, we aggregate this spatially derived question-and-answer information using a robust language model, such as GPT-4o \cite{DBLP:journals/corr/abs-2410-21276}.

\input{boxes/vocot_qry}

\input{boxes/vocot_cmb}

\subsection{SpatialQA}
\label{sec:apx_spaqa}

Unlike Visual-CoT \cite{DBLP:conf/nips/ShaoQ0SZW0024} and VoCoT \cite{DBLP:journals/corr/abs-2405-16919}, the SpatialQA \cite{DBLP:journals/corr/abs-2406-13642} dataset does not provide intermediate CoT reasoning steps or bounding boxes for object identification.
Therefore, we leverage GPT-4o \cite{DBLP:journals/corr/abs-2410-21276}, a powerful multi-modal large language model, to generate detailed synthetic rationale data.
These synthetic rationales supply the necessary intermediate reasoning processes to enable accurate answer generation.

\input{boxes/spatialqa_qry}

\section{LLM-Assistant Evaluation}
\label{sec:llm_eval}

As discussed in Section~\ref{sec:exp}, we utilize the LLM-Assistant evaluation method to assess the data quality of \dataset and measure the performance of \benchmark \cite{DBLP:conf/eccv/LiuDHZZW24, DBLP:conf/nips/LiuLWL23a, DBLP:journals/corr/abs-2503-07265, DBLP:conf/icml/YuYLWL0WW24}.
Evaluation metrics include accuracy and a quantitative score ranging from 0 to 5; both metrics are determined by the LLM-Assistant powered by the Qwen2.5-14B-Instruct-1M \cite{DBLP:journals/corr/abs-2412-15115, DBLP:journals/corr/abs-2501-15383}.

\input{boxes/llm-assistant}

\section{Benchmarks}
\label{sec:bench}

\subsection{Benchmark Employed}
\label{sec:benchs}

We evaluate our method using various benchmarks: SpatialBench \cite{DBLP:journals/corr/abs-2406-13642}, CV-Bench \cite{DBLP:conf/nips/TongBWWIAYYMWPF24}, VSR \cite{}, What's Up \cite{}, our proposed \benchmark, and multiple general VQA benchmarks \cite{}.
Table~\ref{tab:bench} summarizes the statistics for the all benchmarks.
Tables~\ref{tab:main} and \ref{tab:abla} present comparisons between \model and baseline methods, along with comprehensive ablation studies conducted on SpatialBench and \benchmark.
Furthermore, Table~\ref{tab:allbench} summarizes the performance improvements observed across all spatial-related benchmarks.

\input{tables/bench}

\subsection{\benchmark}
\label{sec:bench_con}

To construct the \benchmark dataset, we first filter data samples from \dataset.
Subsequently, we feed each filtered sample into the multi-modal large language models GPT-4o \cite{DBLP:journals/corr/abs-2410-21276} and Gemini-2.5-Pro \cite{DBLP:journals/corr/abs-2403-05530}, using the following prompt to classify the task category.
As mentioned in Section~\ref{sec:exp} and shown in Figure~\ref{fig:bench_pipeline}, if the classification results from both models are consistent, the sample is added to \benchmark; otherwise, it is returned to \dataset.

\input{figures/bench_pipeline}

\input{boxes/bench_cls}

\section{Case Studies}
\label{sec:case_study}

\input{figures/case_study}

To further illustrate the effectiveness of our proposed \model, we provide two example cases in Figure~\ref{fig:case_study}, comparing the performance of \model against five baseline models: Claude-3.5-Haiku \cite{anthropic_claude_3_5_haiku}, GPT-4o-mini \cite{DBLP:journals/corr/abs-2410-21276}, SpatialBot \cite{DBLP:journals/corr/abs-2406-13642}, LLaVA-NeXT-7B \cite{liu2024llavanext}, and the backbone model Qwen2.5-VL-3B \cite{bai2025qwen25vltechnicalreport}.
As shown, our \model consistently produces correct answers, whereas all baseline models fail to provide accurate responses.

In the left example, the images depict only people and bananas.
Consequently, the model must abandon conventional assumptions and carefully reason about the spatial relations explicitly present in the image to answer accurately.
In the right example, complex relationships among numerous objects are depicted, and relevant features for answering the posed question are not immediately obvious.
In this case, the model must thoroughly comprehend the correspondence between each object and the given question, as well as understand intricate spatial relations among these objects, to produce a correct response.
These examples clearly demonstrate that our \model effectively enhances the spatial awareness and reasoning capabilities of vision-language models, thereby significantly improving their ability to understand complex spatial relationships.

%% file: tables/param.tex
\begin{table}[!h]
\caption{Training hyper-parameters of our proposed \model.}
\label{tab:param}
\centering
\resizebox{\columnwidth}{!}{
\begin{tabular}{lcc|lcc}
\toprule
\multicolumn{1}{c}{\bf Configuration} & {\bf Stage 1} & {\bf Stage 2} & \multicolumn{1}{c}{\bf Configuration} & {\bf Stage 1} & {\bf Stage 2} \\
\midrule
Vision Encoder & \multicolumn{2}{c|}{Clip-ViT-Large-Patch14-336 \cite{DBLP:conf/icml/RadfordKHRGASAM21}} & Optimizer & \multicolumn{2}{c}{AdamW \cite{DBLP:conf/iclr/LoshchilovH19}} \\
Depth Encoder & \multicolumn{2}{c|}{Siglip-So400M-Patch14-384 \cite{DBLP:conf/iccv/ZhaiM0B23,DBLP:conf/nips/AlabdulmohsinZ023}} & Learning Rate & \multicolumn{2}{c}{0.00002} \\
Mamba & \multicolumn{2}{c|}{Mamba 130M \cite{gu2024mamba}} & Numerical Precision & \multicolumn{2}{c}{BFloat16} \\
LLM & Qwen2.5 3B \cite{DBLP:journals/corr/abs-2412-15115} & \grey{N/A} & Epoch & 2 & 1 \\
VLM & \grey{N/A} & Qwen2.5-VL 3B \cite{bai2025qwen25vltechnicalreport} & Global Batch Size & 32 & 32 \\
Question Length & \multicolumn{2}{c|}{256} & Learning Schedule & \multicolumn{2}{c}{Cosine Decay} \\
Rational Length & 1024 & \grey{N/A} & Warm-up Ratio & \multicolumn{2}{c}{0.02} \\
Answering Length & \grey{N/A} & 256 & Number of Latent Tokens & \multicolumn{2}{c}{10} \\
\bottomrule
\end{tabular}
}
\end{table}

%% file: tables/ssrbench_score.tex
\begin{table}[!ht]
\caption{Score performance comparison on \benchmark. \benchmark$^{\rm G}$ and \benchmark$^{\rm S}$ denote general and spatial tasks, respectively.}
\label{tab:ssr_score}
\centering
\resizebox{\columnwidth}{!}{
\begin{tabular}{lccllllll}
\toprule
\multicolumn{1}{c}{\multirow{2}{*}{\bf Method}} & \multicolumn{2}{c}{\bf Size} & \multicolumn{3}{c}{\bf \benchmark$^{\rm G}$} & \multicolumn{3}{c}{\bf \benchmark$^{\rm S}$} \\
\cmidrule(lr){2-3} \cmidrule(lr){4-6} \cmidrule(lr){7-9}
 & {\bf Mamba} & {\bf VLM} & \multicolumn{1}{c}{\bf Existence} & \multicolumn{1}{c}{\bf Attribute} & \multicolumn{1}{c}{\bf Action} & \multicolumn{1}{c}{\bf Counting} & \multicolumn{1}{c}{\bf Position} & \multicolumn{1}{c}{\bf Object} \\
\midrule
{\cellcolor[rgb]{ .949,  .949,  .949}\quad\quad \scshape Proprietary} & \multicolumn{8}{c}{\cellcolor[rgb]{ .949,  .949,  .949}} \\
GPT-4o-mini \cite{DBLP:journals/corr/abs-2410-21276} & \grey{N/A} & \grey{N/A} & 4.05 & 2.95 & 3.46 & 3.12 & 2.87 & 2.66 \\
Claude-3.5-Haiku \cite{anthropic_claude_3_5_haiku} & \grey{N/A} & \grey{N/A} & 3.48 & 2.99 & 2.71 & 2.75 & 2.56 & 2.31 \\
\midrule
{\cellcolor[rgb]{ .949,  .949,  .949}\quad\quad \scshape Open-Source} & \multicolumn{8}{c}{\cellcolor[rgb]{ .949,  .949,  .949}} \\
SpatialVLM \cite{DBLP:conf/cvpr/0003XKISGX24, VQASynth} & \grey{N/A} & 3B & 2.34 & 3.32 & 3.55 & 2.34 & 3.56 & 3.24 \\
LLaVA-1.5 \cite{DBLP:conf/cvpr/LiuLLL24} & \grey{N/A} & 7B & 4.17 & 3.72 & 3.66 & 2.71 & 3.87 & 3.56 \\
LLaVA-NeXT \cite{liu2024llavanext} & \grey{N/A} & 7B & 4.23 & 3.59 & 3.79 & 2.66 & 3.69 & 3.41 \\
LLaVA-NeXT \cite{liu2024llavanext} & \grey{N/A} & 13B & 4.30 & 3.82 & 3.79 & 2.76 & 3.78 & 3.12 \\
SpatialBot \cite{DBLP:journals/corr/abs-2406-13642} & \grey{N/A} & 3B & 3.97 & 3.47 & 3.82 & 2.66 & 3.96 & 3.47 \\
Emu3 \cite{DBLP:journals/corr/abs-2409-18869} & \grey{N/A} & 8B & 3.07 & 2.35 & 2.39 & 1.71 & 3.04 & 2.28 \\
Qwen2.5-VL \cite{bai2025qwen25vltechnicalreport} & \grey{N/A} & 3B & 3.56 & 3.42 & 3.56 & 2.41 & 3.43 & 3.00 \\
Qwen2.5-VL \cite{bai2025qwen25vltechnicalreport} & \grey{N/A} & 7B & 4.07 & 3.55 & 3.71 & 2.85 & 3.50 & 3.16 \\
\rowcolor{lightblue}{\bf \model (Ours)} & 130M & 3B & 4.44 & 4.28 & 3.95 & 3.17 & 4.40 & 4.02 \\
\rowcolor{lightblue}{\bf \model (Ours)} & 130M & 7B & 4.65 & 4.17 & 4.10 & 3.71 & 4.43 & 4.16 \\
\bottomrule
\end{tabular}
}
\end{table}

%% file: boxes/viscot_qry.tex
\begin{tcolorbox}[colback=white,colframe=black!75,fonttitle=\bfseries, title=Spatial Query for Visual-CoT,width=\textwidth,breakable]
\begin{enumerate}
    \item What is the object in \bbox? Think step by step, and avoid repetition.
    \item Can you estimate the height and width of \bbox? Think step by step, and avoid repetition.
    \item What is the object to the left of \bbox, and what is its height and width? Think step by step, and avoid repetition.
    \item What is the object to the left of \bbox, and what is its distance to \bbox? Think step by step, and avoid repetition.
    \item What is the object to the right of \bbox, and what is its height and width? Think step by step, and avoid repetition.
    \item What is the object to the right of \bbox, and what is its distance to \bbox? Think step by step, and avoid repetition.
    \item What is the object in front of \bbox, and what is its height and width? Think step by step, and avoid repetition.
    \item What is the object in front of \bbox, and what is its distance to \bbox? Think step by step, and avoid repetition.
    \item What is the object behind \bbox, and what is its height and width? Think step by step, and avoid repetition.
    \item What is the object behind \bbox, and what is its distance to \bbox? Think step by step, and avoid repetition.
    \item What is the object below \bbox, and what is its height and width? Think step by step, and avoid repetition.
    \item What is the object below \bbox, and what is its distance to \bbox? Think step by step, and avoid repetition.
    \item What is the object above \bbox, and what is its height and width? Think step by step, and avoid repetition.
    \item What is the object above \bbox, and what is its distance to \bbox? Think step by step, and avoid repetition.
\end{enumerate}
\end{tcolorbox}

%% file: boxes/viscot_cmb.tex
\begin{tcolorbox}[colback=white,colframe=black!75,fonttitle=\bfseries, title=Rationale Generation for Visual-CoT,width=\textwidth,breakable]
Please generate an image description in continuous paragraphs using these strict guidelines: \\
 \\
Coordinate Usage Rules: \\
1. ONLY use coordinates that are explicitly defined in this mapping table: \\
\begin{varwidth}{\linewidth}\quad - Region [0]: \bbox\end{varwidth} \\
\begin{varwidth}{\linewidth}\quad - ...\end{varwidth} \\
2. Do NOT create or infer any new coordinates \\
3. Each coordinate can only be used ONCE in the description \\
4. Coordinates must be written in [x1,y1,x2,y2] format without spaces \\
 \\
Content Rules: \\
1. Place coordinate immediately after describing its corresponding object \\
2. Integrate coordinates naturally within complete sentences \\
3. Include all provided measurements and spatial relationships \\
4. Maintain narrative flow while incorporating technical details \\
5. Focus on visual elements and their relationships \\
6. Embed coordinates from the mapping table immediately after their corresponding region objects (e.g., "a dog [x1,y1,x2,y2]") \\
7. Maintain paragraph continuity by integrating coordinates within complete sentences \\
8. Preserve strict region-coordinate mapping from the provided table \\
9. Use only [x1,y1,x2,y2] format without spaces \\
10. Exclude technical metadata and region index numbers from final text \\
11. Automatically resolve spatial contradictions using coordinate data \\
12. Ensure coordinate annotations flow naturally after object nouns \\
 \\
Input Data:\\
\brown{Spatial Query and Response for Visual-CoT}
\end{tcolorbox}

%% file: boxes/vocot_qry.tex
\begin{tcolorbox}[colback=white,colframe=black!75,fonttitle=\bfseries, title=Spatial Query for VoCoT,width=\textwidth,breakable]

\textbf{\textit{\underline{Query for each bounding box:}}}

\begin{enumerate}
    \item What is the object in \bbox? Think step by step, and avoid repetition.
    \item Can you estimate the height and width of \bbox? Think step by step, and avoid repetition.
    \item What is the object to the left of \bbox, and what is its height and width? Think step by step, and avoid repetition.
    \item What is the object to the left of \bbox, and what is its distance to \bbox? Think step by step, and avoid repetition.
    \item What is the object to the right of \bbox, and what is its height and width? Think step by step, and avoid repetition.
    \item What is the object to the right of \bbox, and what is its distance to \bbox? Think step by step, and avoid repetition.
    \item What is the object in front of \bbox, and what is its height and width? Think step by step, and avoid repetition.
    \item What is the object in front of \bbox, and what is its distance to \bbox? Think step by step, and avoid repetition.
    \item What is the object behind \bbox, and what is its height and width? Think step by step, and avoid repetition.
    \item What is the object behind \bbox, and what is its distance to \bbox? Think step by step, and avoid repetition.
    \item What is the object below \bbox, and what is its height and width? Think step by step, and avoid repetition.
    \item What is the object below \bbox, and what is its distance to \bbox? Think step by step, and avoid repetition.
    \item What is the object above \bbox, and what is its height and width? Think step by step, and avoid repetition.
    \item What is the object above \bbox, and what is its distance to \bbox? Think step by step, and avoid repetition.
\end{enumerate}

\textbf{\textit{\underline{Query for every two bounding box:}}}

\begin{enumerate}
    \item Which one is higher between \bboxa and \bboxb? Think step by step, and avoid repetition.
    \item Can you estimate how far apart \bboxa and \bboxb are? Think step by step, and avoid repetition.
    \item What direction is \bboxb in relation to \bboxa? Think step by step, and avoid repetition.
    \item How far is \bboxa from \bboxb horizontally? Think step by step, and avoid repetition.
    \item Does \bboxa have a larger size compared to \bboxb? Think step by step, and avoid repetition.
    \item Does \bboxa have a lesser width compared to \bboxb? Think step by step, and avoid repetition.
\end{enumerate}
\end{tcolorbox}

%% file: boxes/vocot_cmb.tex
\begin{tcolorbox}[colback=white,colframe=black!75,fonttitle=\bfseries, title=Rationale Generation for VoCoT,width=\textwidth]
Integrate all measurements values and spatial information from the conversation into answer to get detailed reasoning rationale with spatial details. \\
Then, extract the direct question and answer from question and answer respectively. \\
\\
Content Rules: \\
1. Place coordinate immediately after describing its corresponding object first time, make sure each coordinate appear only once. \\
2. Avoid introducing other coordinates that do not appear in answer. \\
3. Add all provided measurements values and spatial relationships from the conversation to the rationale detailedly. \\
4. Ensure the rationale contains all the information from each sentence in the conversation, especially the measurements values and spatial relationships. \\
5. Automatically resolve spatial contradictions using coordinate data based on the image. \\
\\
Output in the following json template: \\
\{ \\
\begin{varwidth}{\linewidth}\quad "question": <question> \end{varwidth} \\
\begin{varwidth}{\linewidth}\quad , "rationale": <rationale> \end{varwidth} \\
\begin{varwidth}{\linewidth}\quad , "answer": <answer> \end{varwidth} \\
\} \\
\\
Question: \brown{Question} \\
Answer: \brown{Answer} \\
Conversation: \brown{Spatial Query and Response for VoCoT} \\
\end{tcolorbox}

%% file: boxes/spatialqa_qry.tex
\begin{tcolorbox}[colback=white,colframe=black!75,fonttitle=\bfseries, title=Rationale Generation for SpatialQA,width=\textwidth,breakable]
I have an image and a question that I want you to answer. \\
I need you to strictly follow the format with four specific sections: summary, caption, reasoning, and conclusion. \\
It is crucial that you adhere to this structure exactly as outlined and that the final answer in the conclusion matches the standard correct answer precisely. \\
To explain further: \\
\begin{varwidth}{\linewidth}\quad - In summary, briefly explain what steps you'll take to solve the problem.\end{varwidth} \\
\begin{varwidth}{\linewidth}\quad - In caption, describe the contents of the image, specifically focusing on details relevant to the question.\end{varwidth} \\
\begin{varwidth}{\linewidth}\quad - In reasoning, outline a step-by-step thought process you would use to solve the problem based on the image.\end{varwidth} \\
\begin{varwidth}{\linewidth}\quad - In conclusion, give the final answer in a direct format, and it must match the correct answer exactly. If it's a multiple choice question, the conclusion should only include the option without repeating what the option is.\end{varwidth} \\
Finally, integrate these sections into a natural thinking paragraph. \\
 \\
Here's the question and answer: \\
Question: \brown{Question} \\
Answer: \brown{Answer}
\end{tcolorbox}

%% file: boxes/llm-assistant.tex
\begin{tcolorbox}[colback=white,colframe=black!75,fonttitle=\bfseries, title=Prompt for LLM-Assistant VQA Evaluation,width=\textwidth,breakable]
You are an intelligent chatbot designed for evaluating the correctness of generative outputs for question-answer pairs. \\
Your task is to compare the predicted answer with the correct answer and determine if they match meaningfully. Here's how you can accomplish the task: \\
------ \\
\#\#INSTRUCTIONS: \\
- Focus on the meaningful match between the predicted answer and the correct answer. \\
- Consider synonyms or paraphrases as valid matches. \\
- Evaluate the correctness of the prediction compared to the answer. \\
Please evaluate the following image-based question-answer pair: \\
Question: {question} \\
Correct Answer: {answer} \\
Predicted Answer: {response} \\
Provide your evaluation only as a yes/no and score where the score is an integer value between 0 and 5, with 5 indicating the highest meaningful match. \\
Please generate the response in the form of a Python dictionary string with keys 'pred' and 'score', where value of 'pred' is  a string of 'yes' or 'no' and value of 'score' is in INTEGER, not STRING. \\
DO NOT PROVIDE ANY OTHER OUTPUT TEXT OR EXPLANATION. Only provide the Python dictionary string. \\
For example, your response should look like this: \{'pred': 'yes', 'score': 4.8\}.
\end{tcolorbox}

%% file: tables/bench.tex
\begin{table}[!ht]
\caption{Statistics of benchmarks utilized in this paper.}
\label{tab:bench}
\centering
\resizebox{\columnwidth}{!}{
\begin{tabular}{lllclllc}
\toprule
\multicolumn{1}{c}{\bf Benchmark} & \multicolumn{2}{c}{\bf Task} & {\bf Size} & \multicolumn{1}{c}{\bf Benchmark} & \multicolumn{2}{c}{\bf Task} & {\bf Size} \\
\midrule
\multirow{4}{*}{SpatialBench \cite{DBLP:journals/corr/abs-2406-13642}} & \multicolumn{2}{l}{\quad\quad\quad Position} & 34 & \multirow{4}{*}{CV-Bench \cite{DBLP:conf/nips/TongBWWIAYYMWPF24}} & \multicolumn{2}{l}{\quad\quad\quad Count} & 788 \\
 & \multicolumn{2}{l}{\quad\quad\quad Existence} & 40 &  & \multicolumn{2}{l}{\quad\quad\quad Relation} & 650 \\
 & \multicolumn{2}{l}{\quad\quad\quad Counting} & 20 &  & \multicolumn{2}{l}{\quad\quad\quad Depth} & 600 \\
 & \multicolumn{2}{l}{\quad\quad\quad Size} & 40 &  & \multicolumn{2}{l}{\quad\quad\quad Distance} & 600 \\
\cmidrule(lr){2-4} \cmidrule(lr){6-8}
\multicolumn{3}{r}{\cellcolor[rgb]{ .949,  .949,  .949}\textsc{Total}} & \cellcolor[rgb]{ .949,  .949,  .949}\textit{114} & \multicolumn{3}{r}{\cellcolor[rgb]{ .949,  .949,  .949}\textsc{Total}} & \cellcolor[rgb]{ .949,  .949,  .949}\textit{2638} \\
\midrule
\multirow{4}{*}{\benchmark} & \multirow{3}{*}{General} & Existence & 107 &  & \multirow{3}{*}{Spatial} & Counting & 41 \\
 & & Attribute & 168 & & & Position & 162 \\
 & & Action & 157 & &  & Object & 154 \\
\cmidrule(lr){2-8}
\multicolumn{7}{r}{\cellcolor[rgb]{ .949,  .949,  .949}\textsc{Total}} & \cellcolor[rgb]{ .949,  .949,  .949}\textit{789} \\
\midrule
\multirow{2}{*}{VSR \cite{DBLP:journals/tacl/0001EC23}} & \multicolumn{2}{l}{\quad\quad\quad Random} & 1222 & & \multicolumn{2}{l}{\quad\quad\quad Zero-Shot} & 2195 \\
\cmidrule(lr){2-8}
\multicolumn{7}{r}{\cellcolor[rgb]{ .949,  .949,  .949}\textsc{Total}} & \cellcolor[rgb]{ .949,  .949,  .949}\textit{3417} \\
\midrule
What's Up \cite{DBLP:conf/emnlp/KamathHC23a} & \multicolumn{2}{r}{\cellcolor[rgb]{ .949,  .949,  .949}\textsc{Total}} & \cellcolor[rgb]{ .949,  .949,  .949}\textit{820} & VQAv2 \cite{DBLP:conf/cvpr/GoyalKSBP17} & \multicolumn{2}{r}{\cellcolor[rgb]{ .949,  .949,  .949}\textsc{Total}} & \cellcolor[rgb]{ .949,  .949,  .949}\textit{107k} \\
TextVQA \cite{DBLP:conf/cvpr/SinghNSJCBPR19} & \multicolumn{2}{r}{\cellcolor[rgb]{ .949,  .949,  .949}\textsc{Total}} & \cellcolor[rgb]{ .949,  .949,  .949}\textit{5k} & POPE \cite{DBLP:conf/emnlp/LiDZWZW23} & \multicolumn{2}{r}{\cellcolor[rgb]{ .949,  .949,  .949}\textsc{Total}} & \cellcolor[rgb]{ .949,  .949,  .949}\textit{9k} \\
MMBench \cite{DBLP:conf/eccv/LiuDZLZZYWHLCL24} & \multicolumn{2}{r}{\cellcolor[rgb]{ .949,  .949,  .949}\textsc{Total}} & \cellcolor[rgb]{ .949,  .949,  .949}\textit{3k} & GQA \cite{DBLP:conf/cvpr/HudsonM19} & \multicolumn{2}{r}{\cellcolor[rgb]{ .949,  .949,  .949}\textsc{Total}} & \cellcolor[rgb]{ .949,  .949,  .949}\textit{12k} \\
\bottomrule
\end{tabular}
}
\end{table}

%% file: figures/bench_pipeline.tex
\begin{figure}[!ht]
    \centering
    \includegraphics[width=\linewidth]{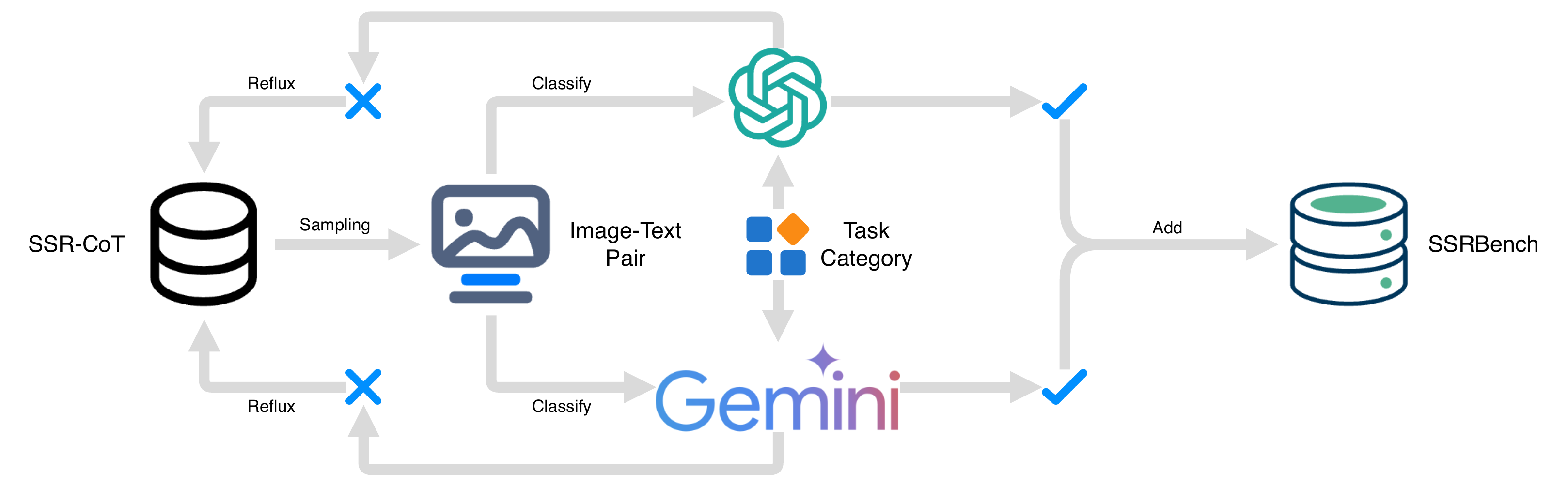}
    \caption{Schematic of \benchmark construction pipeline.}
    \label{fig:bench_pipeline}
\end{figure}

%% file: boxes/bench_cls.tex
\begin{tcolorbox}[colback=white,colframe=black!75,fonttitle=\bfseries, title=Task classification for \benchmark,width=\textwidth,breakable]
You are an expert in image-based question classification. \\
You need to classify each input question into a specific task type based on the following taxonomy. \\
Task Categories: \\
\begin{varwidth}{\linewidth}\quad \green{Spatial}:\end{varwidth} \\
\begin{varwidth}{\linewidth}\quad\quad Explanation: Involve identifying and understanding the position, size, shape, and relative relationships of objects in an image.\end{varwidth} \\
\begin{varwidth}{\linewidth}\quad\quad Subtasks:\end{varwidth} \\
\begin{varwidth}{\linewidth}\quad\quad\quad \blue{Count}: Counting objects in the image (e.g., questions like “How many ...?”).\end{varwidth} \\
\begin{varwidth}{\linewidth}\quad\quad\quad \blue{Relative Position Recognition}: Determining spatial relations like “to the left of”, “above”, or “on the right”.\end{varwidth} \\
\begin{varwidth}{\linewidth}\quad\quad\quad \blue{Position Based Object Recognition}: Identifying an object based on its spatial relation to another object (e.g., “What is the object to the left of the dog?”).\end{varwidth} \\
\begin{varwidth}{\linewidth}\quad \green{General}:\end{varwidth} \\
\begin{varwidth}{\linewidth}\quad\quad Explanation: Involve classifying, recognizing, or reasoning about visual content without necessarily focusing on spatial relations.\end{varwidth} \\
\begin{varwidth}{\linewidth}\quad\quad Subtasks:\end{varwidth} \\
\begin{varwidth}{\linewidth}\quad\quad\quad \blue{Existence}: Determining whether an object or feature is present (e.g., “Is there a cat?”).\end{varwidth} \\
\begin{varwidth}{\linewidth}\quad\quad\quad \blue{Attribute Recognition}: Identifying attributes like color, texture, size, or state (e.g., “What color is the apple?”).\end{varwidth} \\
\begin{varwidth}{\linewidth}\quad\quad\quad \blue{Action Recognition}: Recognizing what action or activity is occurring (e.g., “What is the man doing?”).\end{varwidth} \\
 \\
For each input question: \\
\begin{varwidth}{\linewidth}\quad First determine whether the question belongs to the spatial or general category.\end{varwidth} \\
\begin{varwidth}{\linewidth}\quad Then classify it into one of the three subtasks under that category.\end{varwidth} \\
\begin{varwidth}{\linewidth}\quad If the question does not match any of the subtasks under either category, return None.\end{varwidth} \\
 \\
Output format: \\
\begin{varwidth}{\linewidth}\quad \{"category": "spatial" or "general", "subtask": "subtask\_name" or "None"\}\end{varwidth} \\
 \\
Example Input: "Is there a bicycle in the image?" \\
Example Output: \{"category": "general", "subtask": "existence"\} \\
 \\
Now, let's begin classification. Here's the question: \\
Question: \brown{Question}
\end{tcolorbox}

%% file: figures/case_study.tex
\begin{figure}[!ht]
    \centering
    \includegraphics[width=\linewidth]{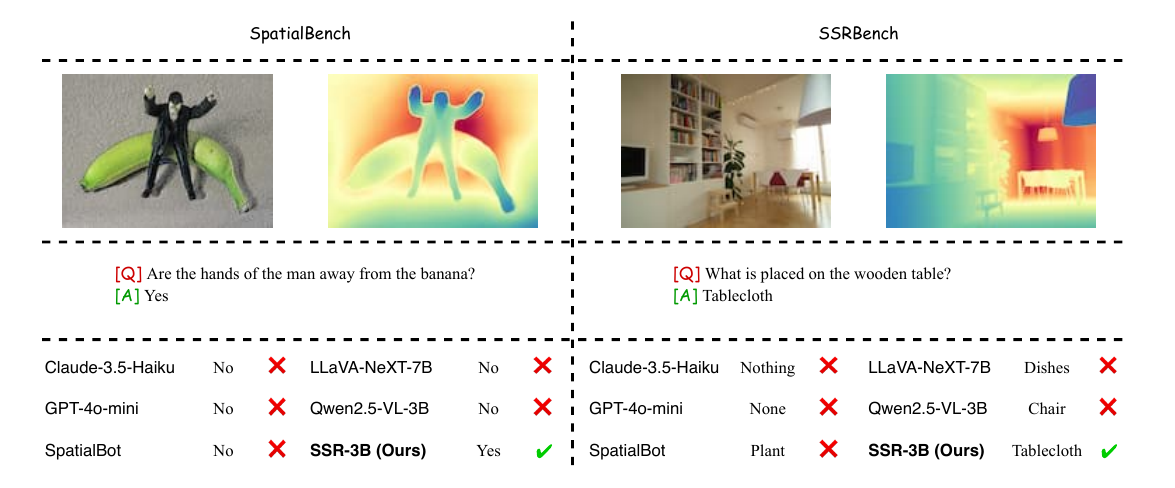}
    \caption{Two examples illustrating question-answering performance by baseline models compared to our \model are presented.}
    \label{fig:case_study}
\end{figure}

%% file: sections/8_checklist.tex
\clearpage
\newpage
\section*{NeurIPS Paper Checklist}

\begin{enumerate}

\item {\bf Claims}
    \item[] Question: Do the main claims made in the abstract and introduction accurately reflect the paper's contributions and scope?
    \item[] Answer: \answerYes{} 
    \item[] Justification: The claims we presented in the abstract and introduction are clearly stated and fully aligned with the contributions of this paper.

\item {\bf Limitations}
    \item[] Question: Does the paper discuss the limitations of the work performed by the authors?
    \item[] Answer: \answerYes{} 
    \item[] Justification: We discuss the limitations in the Section~\ref{sec:con}..

\item {\bf Theory assumptions and proofs}
    \item[] Question: For each theoretical result, does the paper provide the full set of assumptions and a complete (and correct) proof?
    \item[] Answer: \answerNA{} 
    \item[] Justification: None of theoretical assumptions.

    \item {\bf Experimental result reproducibility}
    \item[] Question: Does the paper fully disclose all the information needed to reproduce the main experimental results of the paper to the extent that it affects the main claims and/or conclusions of the paper (regardless of whether the code and data are provided or not)?
    \item[] Answer: \answerYes{} 
    \item[] Justification: We provide the implementation details to reproduce the main experimental results in Section~\ref{sec:exp}.

\item {\bf Open access to data and code}
    \item[] Question: Does the paper provide open access to the data and code, with sufficient instructions to faithfully reproduce the main experimental results, as described in supplemental material?
    \item[] Answer: \answerYes{} 
    \item[] Justification: Instructions for reproducing our experiments are provided in Section~\ref{sec:exp}. We will publicly release the data and code once they have been finalized and prepared.
    \item[] Guidelines:
    \begin{itemize}
        \item The answer NA means that paper does not include experiments requiring code.
        \item Please see the NeurIPS code and data submission guidelines (\url{https://nips.cc/public/guides/CodeSubmissionPolicy}) for more details.
        \item While we encourage the release of code and data, we understand that this might not be possible, so “No” is an acceptable answer. Papers cannot be rejected simply for not including code, unless this is central to the contribution (e.g., for a new open-source benchmark).
        \item The instructions should contain the exact command and environment needed to run to reproduce the results. See the NeurIPS code and data submission guidelines (\url{https://nips.cc/public/guides/CodeSubmissionPolicy}) for more details.
        \item The authors should provide instructions on data access and preparation, including how to access the raw data, preprocessed data, intermediate data, and generated data etc.
        \item The authors should provide scripts to reproduce all experimental results for the new proposed method and baselines. If only a subset of experiments are reproducible, they should state which ones are omitted from the script and why.
        \item At submission time, to preserve anonymity, the authors should release anonymized versions (if applicable).
        \item Providing as much information as possible in supplemental material (appended to the paper) is recommended, but including URLs to data and code is permitted.
    \end{itemize}

\item {\bf Experimental setting/details}
    \item[] Question: Does the paper specify all the training and test details (e.g., data splits, hyperparameters, how they were chosen, type of optimizer, etc.) necessary to understand the results?
    \item[] Answer: \answerYes{} 
    \item[] Justification: The experimental setup required for our study is described in Section~\ref{sec:exp}.

\item {\bf Experiment statistical significance}
    \item[] Question: Does the paper report error bars suitably and correctly defined, or other appropriate information about the statistical significance of the experiments?
    \item[] Answer: \answerNo{} 
    \item[] Justification: Conducting experimentfor  statistical significance when training large-scale models typically demands an exponential increase in computational resources. To address this concern, we performed comprehensive multi-angle analyses on \model, and our experimental results consistently validated its effectiveness.
    \item[] Guidelines:
    \begin{itemize}
        \item The answer NA means that the paper does not include experiments.
        \item The authors should answer "Yes" if the results are accompanied by error bars, confidence intervals, or statistical significance tests, at least for the experiments that support the main claims of the paper.
        \item The factors of variability that the error bars are capturing should be clearly stated (for example, train/test split, initialization, random drawing of some parameter, or overall run with given experimental conditions).
        \item The method for calculating the error bars should be explained (closed form formula, call to a library function, bootstrap, etc.)
        \item The assumptions made should be given (e.g., normally distributed errors).
        \item It should be clear whether the error bar is the standard deviation or the standard error of the mean.
        \item It is OK to report 1-sigma error bars, but one should state it. The authors should preferably report a 2-sigma error bar rather than state that they have a 96\% CI, if the hypothesis of Normality of errors is not verified.
        \item For asymmetric distributions, the authors should be careful not to show in tables or figures symmetric error bars that would yield results that are out of range (e.g., negative error rates).
        \item If error bars are reported in tables or plots, the authors should explain in the text how they were calculated and reference the corresponding figures or tables in the text.
    \end{itemize}

\item {\bf Experiments compute resources}
    \item[] Question: For each experiment, does the paper provide sufficient information on the computer resources (type of compute workers, memory, time of execution) needed to reproduce the experiments?
    \item[] Answer: \answerYes{} 
    \item[] Justification: Details regarding the computational resources used for all experiments are presented in Section~\ref{sec:exp}.
    \item[] Guidelines:
    \begin{itemize}
        \item The answer NA means that the paper does not include experiments.
        \item The paper should indicate the type of compute workers, CPU or GPU, internal cluster, or cloud provider, including relevant memory and storage.
        \item The paper should provide the amount of compute required for each of the individual experimental runs, as well as estimate the total compute. 
        \item The paper should disclose whether the full research project required more computing than the experiments reported in the paper (e.g., preliminary or failed experiments that didn't make it into the paper). 
    \end{itemize}
    
\item {\bf Code of ethics}
    \item[] Question: Does the research conducted in the paper conform, in every respect, with the NeurIPS Code of Ethics \url{https://neurips.cc/public/EthicsGuidelines}?
    \item[] Answer: \answerYes{} 
    \item[] Justification: The research presented in this paper fully complies with the NeurIPS Code of Ethics.

\item {\bf Broader impacts}
    \item[] Question: Does the paper discuss both potential positive societal impacts and negative societal impacts of the work performed?
    \item[] Answer: \answerYes{} 
    \item[] Justification: We discuss impacts in the Section~\ref{sec:con}.
    
\item {\bf Safeguards}
    \item[] Question: Does the paper describe safeguards that have been put in place for the responsible release of data or models that have a high risk for misuse (e.g., pretrained language models, image generators, or scraped datasets)?
    \item[] Answer: \answerNA{} 
    \item[] Justification: We used a publicly available visual instruction tuning dataset and pre-trained visual foundation models and large language models.
    \item[] Guidelines:
    \begin{itemize}
        \item The answer NA means that the paper poses no such risks.
        \item Released models that have a high risk for misuse or dual-use should be released with necessary safeguards to allow for controlled use of the model, for example, by requiring that users adhere to usage guidelines or restrictions to access the model or implementing safety filters. 
        \item Datasets that have been scraped from the Internet could pose safety risks. The authors should describe how they avoided releasing unsafe images.
        \item We recognize that providing effective safeguards is challenging, and many papers do not require this, but we encourage authors to take this into account and make a best-faith effort.
    \end{itemize}

\item {\bf Licenses for existing assets}
    \item[] Question: Are the creators or original owners of assets (e.g., code, data, models) used in the paper properly credited, and are the license and terms of use explicitly mentioned and properly respected?
    \item[] Answer: \answerYes{} 
    \item[] Justification: We cited models and datasets we deal with in this paper.

\item {\bf New assets}
    \item[] Question: Are new assets introduced in the paper well documented, and is the documentation provided alongside the assets?
    \item[] Answer: \answerYes{} 
    \item[] Justification: We propose a novel model \model, curate a dataset \dataset, and construct a benchmark \benchmark.

\item {\bf Crowdsourcing and research with human subjects}
    \item[] Question: For crowdsourcing experiments and research with human subjects, does the paper include the full text of instructions given to participants and screenshots, if applicable, as well as details about compensation (if any)? 
    \item[] Answer: \answerNA{} 
    \item[] Justification: We do not include human subjects in this paper.

\item {\bf Institutional review board (IRB) approvals or equivalent for research with human subjects}
    \item[] Question: Does the paper describe potential risks incurred by study participants, whether such risks were disclosed to the subjects, and whether Institutional Review Board (IRB) approvals (or an equivalent approval/review based on the requirements of your country or institution) were obtained?
    \item[] Answer: \answerNA{} 
    \item[] Justification: We do not include human subjects in this paper.

\item {\bf Declaration of LLM usage}
    \item[] Question: Does the paper describe the usage of LLMs if it is an important, original, or non-standard component of the core methods in this research? Note that if the LLM is used only for writing, editing, or formatting purposes and does not impact the core methodology, scientific rigor, or originality of the research, the declaration is not required.
    \item[] Answer: \answerYes{} 
    \item[] Justification: We utilized LLM for data synthesis and the reporting of performance indicators, which are comprehensively described throughout this paper.

\end{enumerate}